\documentclass{new_tlp}

\usepackage[utf8]{inputenc}
\usepackage{microtype}
\usepackage{amsmath,amssymb}
\usepackage{tikz}
\usetikzlibrary{shapes,arrows}
\usetikzlibrary{calc}
\usepackage[linesnumbered]{algorithm2e}
\usepackage{times,helvet,courier}
\usepackage{url}
\usepackage{listings}
\usepackage{mathtools}

\newcommand{\sysfont}{\textit}
\newcommand{\clingo}{\sysfont{clingo}}
\newcommand{\clingoDL}{\clingo{\small\textnormal{[}\textsc{DL}\textnormal{]}}}
\newcommand{\code}[1]{\texttt{#1}}
\newcommand{\bind}[2]{\ensuremath{\mathit{Bin}_{#1,#2}}}
\newcommand{\lind}[3]{\ensuremath{\mathit{Lin}^{#3}_{#1,#2}}}

\newcommand{\vbs}{\textsc{vbs}}
\newcommand{\runtime}{\textsc{t}}
\newcommand{\grounding}{\textsc{gt}}
\newcommand{\quality}{\textsc{qu}}
\newcommand{\aquality}{\textsc{aqu}}
\newcommand{\choices}{\textsc{ch}}
\newcommand{\conflicts}{\textsc{co}}
\newcommand{\conf}{\textsc{conf}}
\newcommand{\instance}{\textsc{ins}}

\newcommand{\trains}{\textsc{\#tl}}
\newcommand{\resources}{\textsc{\#r}}
\newcommand{\subsumed}{\textsc{\#sr}}
\newcommand{\resourcextl}{\textsc{\#rtl}}
\newcommand{\resourceareas}{\textsc{\#ra}}
\newcommand{\edgeconflicts}{\textsc{\#ec}}
\newcommand{\areaconflicts}{\textsc{\#rac}}
\newcommand{\nodenaming}{\textsc{\#vnn}}
\newcommand{\toponaming}{\textsc{\#vhn}}

\newcommand{\none}{\textsc{def}}
\newcommand{\hseq}{\textsc{hs}}

\newcommand{\ooone}{\textsc{ol}{\scriptsize 1}}
\newcommand{\ootwo}{\textsc{ol}{\scriptsize 2}}
\newcommand{\ot}{\textsc{ac}}

\allowdisplaybreaks

\title[Train Scheduling with Hybrid ASP]{Train Scheduling with Hybrid Answer Set Programming%
  \thanks{This is a substantially extended and revised version of~({Abels et~al\mbox{.}}~{2019}).}%
  \thanks{This work was partially funded by DFG grants SCHA 550/9 and~11.}}

\author[Dirk Abels et al.]{%
  DIRK ABELS, JULIAN JORDI \\ SBB, Switzerland
  \and
  MAX OSTROWSKI \\ Potassco Solutions, Germany
  \and
  TORSTEN SCHAUB\thanks{Affiliated with Simon Fraser University, Canada, and Griffith University, Australia.}\\
  Potassco Solutions, Germany, and University of Potsdam, Germany
  \and
  AMBRA TOLETTI \\ SBB, Switzerland
  \and
  PHILIPP WANKO \\ Potassco Solutions, Germany, and University of Potsdam, Germany
  }

\begin{document}

\maketitle

\begin{abstract}
  We present a solution to real-world train scheduling problems,
  involving routing, scheduling, and optimization,
  based on Answer Set Programming (ASP).
  To this end, we pursue a hybrid approach that extends ASP with
  difference constraints to account for a fine-grained timing.
  More precisely, we exemplarily show how the hybrid ASP system \clingoDL{}
  can be used to tackle demanding planning-and-scheduling problems.
  In particular, we investigate how to boost performance by combining distinct ASP solving techniques,
  such as approximations and heuristics,
  with preprocessing and encoding techniques for tackling large-scale, real-world train scheduling instances.
\end{abstract}
%

\section{Introduction}\label{sec:introduction}

Densely-populated railway networks transport millions of people and carry millions of tons of freight daily;
and this traffic is expected to increase even further.
Hence, for using a railway network to capacity,
it is important to schedule trains in a flexible and global way.
This is however far from easy since the generation of railway timetables is already known to be intractable for a single track~\cite{cafito02a}.
Moreover,
hundreds of train lines on a densely connected railway network lead to complex inter-dependencies due to connections between trains and resource conflicts.

We take up this challenge and show how to address real-world train scheduling with hybrid Answer Set Programming (ASP~\cite{lifschitz99b}).
Our hybrid approach allows us to specifically account for the different types of constraints induced by routing, scheduling, and optimization.
While we address paths and conflicts with regular ASP,
we use difference constraints (over integers) to capture fine timings.
Similarly, to boost (multi-objective) optimization, we study approximations of delay functions of varying granularity.
This is complemented by domain-specific heuristics aiming at improving feasibility checking.
Moreover, we introduce preprocessing techniques to reduce the problem size and search space,
and provide redundant constraints improving propagation.
This lifts our approach to a level that allows us to create high quality train schedules spanning six hours for a portion of Switzerland within minutes.

In current practice, train schedules evolve by only minor adjustments year-over-year.
The basic structure of the schedule has been shown to be more or less feasible in practice.
The current scheduling tools offer limited support for conflict detection and no support for automatic conflict resolution.
When additional trains are ordered by railway companies during the year, or when capacity is reduced because of maintenance work, it is up to the
professional experience and ingenuity of the planner to find a producible schedule or to reduce the number of services if no feasible solution can be
found.
In a first step, the solution described in this paper are intended to be used by the industrial partner in a decision support system providing
planners with conflict-free schedules for adding additional trains into an existing schedule structure.
The planner only has to enter the commercial requirements for the additional train, everything else shall be taken care of by the system.
In later steps, it is hoped that our solution can be scaled to eventually generate complete, conflict free schedules from scratch for the whole
country and also continually optimize them during operations to account for deviations and disruptions.

We implement our approach with the hybrid ASP system \clingoDL~\cite{jakaosscscwa17a},
an extension of \clingo~\cite{gekakasc17a} with difference constraints.
To begin with,
we introduce in Section~\ref{sec:application} a dedicated formalization of the train scheduling problem.
This is indispensable to master the complex inter-dependencies of the problem.
We present our solution in terms of hybrid ASP encodings,
including a detailed description of preprocessing, encoding and heuristic techniques in Section~\ref{sec:approach}.
Finally, we evaluate our approach on real-word instances in Section~\ref{sec:experiments}.


\section{Background}\label{sec:background}
We rely on a basic acquaintance with ASP.
The syntax of our logic programs follows the one of \clingo~\cite{PotasscoUserGuide};
its semantics is detailed in~\cite{gehakalisc15a}.

The system \clingoDL{} extends the input language of \clingo{} by (theory) atoms representing \emph{difference constraints}.
That is, atoms of the form
\begin{lstlisting}[mathescape,numbers=none]
   &diff{$\,u$-$v\,$} <= $d$
\end{lstlisting}
where $u,v$ are symbolic terms and $d$ is a numeral term,
represent difference constraints such as `$u-v \leq d$',
where $u,v$ serve as integer variables and $d$ stands for an integer.%
\footnote{Strictly speaking, we had to distinguish the integer from its representation.}
For instance,
assume that
`\lstinline[mathescape]|&diff{e(T)-b(T)} <= D|'
stands for the condition that the time between the end and the beginning of a task \texttt{T} must be less or equal than some duration $D$.
This may get instantiated to
`\lstinline[mathescape]|&diff{e(7)-b(7)} <= 42|'
to require that \texttt{e(7)} and \texttt{b(7)} take integer values such that `$\mathtt{e(7)}-\mathtt{b(7)} \leq 42$'.
Note that $u,v$ can be arbitrary terms.
We exploit this below and use instances of pairs like \texttt{(T,N)} to denote integer variables.
Among the alternative semantic couplings between (theory) atoms and constraints offered by \clingoDL{} (cf.\ \cite{jakaosscscwa17a,gekakaosscwa16a}),
we follow the \emph{defined, non-strict} approach
(i) tolerating theory atoms in rule heads and
(ii) enforcing their corresponding constraints only if the atoms are derivable.
Hence, if a theory atom is false, its associated constraint is ignored.
This approach has the advantage that we only need to consider difference constraints occurring in the encoding and not their complements.
Obviously, one great benefit of using such constraints is that their variables are not subject to grounding.

For boosting performance, we take advantage of \clingo's heuristic directives of form
\[
\code{\#heuristic $a$:$B$.$\,$[$w$,sign]}
\]
where $a$ is an atom and $B$ is a body; $w$ is a numeral term and \texttt{sign} a heuristic modifier,
indicating how the solver's heuristic treatment of $a$ should be changed whenever $B$ holds.
Whenever $a$ is chosen by the solver,
\texttt{sign} enforces that it becomes either true or false depending on whether $w$ is positive or negative, respectively.
See~\cite{gekaotroscwa13a,PotasscoUserGuide} for a comprehensive introduction to heuristic modifiers in \clingo.


\section{Real-world train scheduling}\label{sec:application}

\subsection{Problem introduction}

The train scheduling problem can essentially be divided into three distinct tasks: routing, conflict resolution and scheduling.

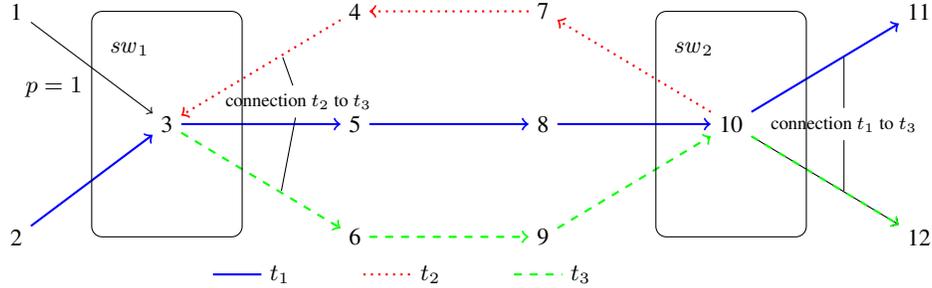
\begin{figure}[ht]
{\begin{tikzpicture}
\tikzstyle{t1sg}=[->]
\tikzstyle{t2sg}=[->,dotted]
\tikzstyle{t3sg}=[->,dashed]
\tikzstyle{t1}=[blue,thick]
\tikzstyle{t2}=[red,thick]
\tikzstyle{t3}=[green,thick]

\node (v1) at (-4,2) {1};
\node at (-3.5,1){$p=1$};
\node (v3) at (-4,-1) {2};
\node (v2) at (-2,0.5) {3};
\node (v4) at (0.5,0.5) {5};
\node (v5) at (3,0.5) {8};
\node (v6) at (5.5,0.5) {10};
\node (v8) at (8,2) {11};
\node (v7) at (8,-1) {12};
\node (v10) at (0.5,2) {4};
\node (v9) at (3,2) {7};
\node (v11) at (0.5,-1) {6};
\node (v12) at (3,-1) {9};
\draw[t1sg]  (v1) edge (v2);
\draw[t1sg,t1]  (v3) edge (v2);
\draw[t1sg,t1]  (v2) edge (v4);
\draw[t1sg,t1]  (v4) edge (v5);
\draw[t1sg,t1]  (v5) edge (v6);
\draw[t1sg]  (v6) edge (v7);
\draw[t1sg,t1]  (v6) edge (v8);
\draw[t2sg,t2]  (v6) edge (v9);
\draw [t2sg,t2] (v9) edge (v10);
\draw [t2sg,t2] (v10) edge (v2);
\draw [t3sg, t3] (v2) edge (v11);
\draw [t3sg, t3] (v11) edge (v12);
\draw [t3sg, t3] (v12) edge (v6);
\draw [t3sg, t3] (v6) edge (v7);
\node (v14) at (-0.5,1.5) {};
\node[font=\scriptsize] (v13) at (-0.25,.8) {connection $t_2$ to $t_3$};
\draw  (v13) edge (v14);
\node (v15) at (7,1.5) {};
\node (v17) at (7,-0.5) {};
\node[font=\scriptsize]  (v16) at (7,0.5) {connection $t_1$ to $t_3$};
\draw  (v15) edge (v16);
\draw  (v17) edge (v16);
\draw[rounded corners]  (-3,2) rectangle (-1,-1);
\draw[rounded corners]  (4.5,2) rectangle (6.5,-1);
\node at (-2.5,1.5) {$
\mathit{sw}_1$};
\node at (5,1.5) {$
\mathit{sw}_2$};
\node (v18) at (-0.5,-0.5) {};
\draw  (v18) edge (v13);
\node (v19) at (-1.5,-1.5) {};
\node (v20) at (-0.5,-1.5) {$t_1$};
\node (v21) at (0.5,-1.5) {};
\node (v22) at (1.5,-1.5) {$t_2$};
\node (v23) at (2.5,-1.5) {};
\node (v24) at (3.5,-1.5) {$t_3$};
\draw [t1] (v19) edge (v20);
\draw [dotted,t2] (v21) edge (v22);
\draw [dashed, t3] (v23) edge (v24);
\end{tikzpicture}}
\caption{Routing of three train lines through a railway network.\label{fig:paths}}
\end{figure}
%
First, train lines are routed through a railway network.
One such network is given in Figure~\ref{fig:paths}.
By and large, it is a directed graph containing nodes 1 through 12 with edges in between, for instance, $(2,3)$ and $(10,12)$.
Given this directed graph, we assign paths through the network to three train lines, viz.\ $t_1$, $t_2$ and $t_3$.
Each train line is assigned an acyclic subgraph capturing its travel trajectory.
In our example,
the subgraphs for $t_1$, $t_2$ and $t_3$ are indicated by
solid, dotted and dashed edges, respectively.
Note that $(10,12)$ belongs to the subgraph of both $t_1$ and $t_3$.
We see that $t_1$ has several different start nodes, viz.\ 1 and 2, and end nodes, viz.\ 11 and 12,
whereas $t_2$ and $t_3$ have no alternative routes.
One possible solution to the routing task in Figure~\ref{fig:paths} is represented by edges colored blue, red and green
marking valid paths for $t_1$, $t_2$ and $t_3$, respectively

Second, edges in the network are assigned resources representing,
for example, the physical tracks or junctions that can only be passed by a single train at once.
Whenever two train lines share edges assigned the same resource,
a decision has to be made which train line passes first.
In our example,
we assume that each edge is assigned a resource representing the physical track.
Furthermore, we have two junctions, $\mathit{sw}_1$ and $\mathit{sw}_2$, that are assigned five edges each.
More precisely, $\mathit{sw}_1$ is assigned to $(1,3),(2,3),(3,5),(3,6)$ and $(4,3)$,
and $\mathit{sw}_2$ to $(8,10),(9,10),(10,11),(10,12)$ and $(10,7)$.
The junctions highlight a common structural property of train scheduling instances
on which we rely heavily to reduce the amount of decisions that have to be made to resolve resource conflicts.
Let us focus on $\mathit{sw}_2$ and the resource conflicts of the three train lines in edges $(8,10),(9,10),(10,7),(10,11)$ and $(10,12)$.
Instead of deciding each pair of edges with shared resources individually,
we can decide which train line enters a set of edges assigned $\mathit{sw}_2$ first.
More precisely, we decide in which order $t_1$, $t_2$ and $t_3$ enter $\{(8,10),(10,11),(10,12)\}$, $\{(10,7)\}$ and $\{(9,10),(10,12)\}$, respectively.
This is possible because train lines using the same resource cannot overtake each other once they are within these sets of edges.
We call such sets \emph{resource areas}.
Note that while it is obvious in our example what said areas are
(they are equal to the full set of assigned edges for each resource and train line),
this may not be the case in general.
For instance,
there are many alternative paths in complex train stations,
and thus shared resources might be visited several times
enabling train lines to overtake one another.

Finally, a schedule has to be created that assigns each train line an arrival time at all nodes in its path.
A valid schedule has to respect a variety of timing constraints,
ranging from earliest and latest arrival times at nodes,
traveling and waiting times on edges,
resource conflicts between train lines,
to connections between train lines.
\begin{figure}[ht]
\centering
{\begin{tikzpicture}
\tikzstyle{t1}=[blue,thick]
\tikzstyle{t2}=[red,thick]
\tikzstyle{t3}=[green,thick]
\node (v1) at (-0.5,0) {};
\node[blue,anchor=south] at (.5,1) {$t_1$};
\node[blue,anchor=south] (v5) at (3,1) {2};
\node[blue,anchor=south] at (4,1) {3};
\node[blue,anchor=south] at (5,1) {5};
\node[blue,anchor=south] at (6,1) {8};
\node[blue,anchor=south] at (7,1) {10};
\node[blue,anchor=south] at (8,1) {11};
\node[red,anchor=south] at (.5,.5) {$t_2$};
\node[red,anchor=south] at (1,.5) {10};
\node[red,anchor=south] at (2,.5) {7};
\node[red,anchor=south] at (3,.5) {4};
\node[red,anchor=south] at (4,.5) {3};
\node[green,anchor=south] at (.5,0) {$t_3$};
\node[green,anchor=south] at (4,0) {3};
\node[green,anchor=south] at (5,0) {6};
\node[green,anchor=south] at (6,0) {9};
\node[green,anchor=south] at (7,0) {10};
\node[green,anchor=south] at (8,0) {12};
\node (v2) at (8.5,0) {};
\draw[->] (v1) -- (v2) node[anchor=west] {$V$};
    	\foreach \x in {0,...,8}
     		\draw (\x,2pt) -- (\x,-2pt);
\node (v3) at (0,0.5) {};
\node (v4) at (0,-7) {};
\draw[->] (v3) -- (v4) node[anchor=north] {$t$};
    	\foreach \y in {6,...,1}
     		\draw (1pt,-\y) -- (-3pt,-\y) 
     			node[anchor=east] {%
    \pgfmathparse{120*\y}
    \pgfmathprintnumber[    
        fixed,
        fixed zerofill,
        precision=0
    ]{\pgfmathresult}%
    }; 
\draw [help lines, step=1cm,opacity=.3] (0,-7) grid (8,0);
\draw[blue,dashed] (0.5,-2.5) -- (8.5,-6.5)node[anchor=south]{$d_{t_1}$};
\draw[red,dashed] (0.5,-1.75)node[anchor=south]{$d_{t_2}$} -- (4,-3.5);
\draw[green,dashed] (4,-3.5) -- (8.5,-5.75)node[anchor=south]{$d_{t_3}$};
\fill[red,opacity=.2] (1,0) node (v7) {} -- (2,-0.5) -- (3,-1) -- (4,-1.5) node (v6){}-- (4,-4.5)-- (3,-4)-- (2,-3.5)-- (1,-3) ;

\draw [t2](1,0) node (v7) {} -- (2,-0.5) -- (3,-1) -- (4,-1.5) node (v6) {};
\draw [t3](4,-1.5) -- (5,-2) -- (6,-5.5) -- (7,-6) -- (8,-6.5);

\draw [t1](3,-2.5) -- (4,-3) -- (5,-3.5) -- (6,-4) -- (7,-4.5) -- (8,-5);
\fill[green,opacity=.2] (4,-1.5) node (v8) {} -- (5,-2) -- (6,-2.5) -- (7,-3) -- (8,-3.5) -- (8,-6.5) -- (7,-6) -- (6,-5.5) -- (5,-5) -- (4,-4.5) --(4,-1.5);
\fill[blue,opacity=.2] (3,-2) node (v9) {} -- (4,-2.5) -- (5,-3) -- (6,-3.5) -- (7,-4) -- (8,-4.5) -- (8,-7) -- (7,-6.5) -- (6,-6) -- (5,-5.5) -- (4,-5) -- (3,-4.5) -- (3,-2);
\end{tikzpicture}}%
\caption{Scheduling of three train lines.\label{fig:times}}
\end{figure}
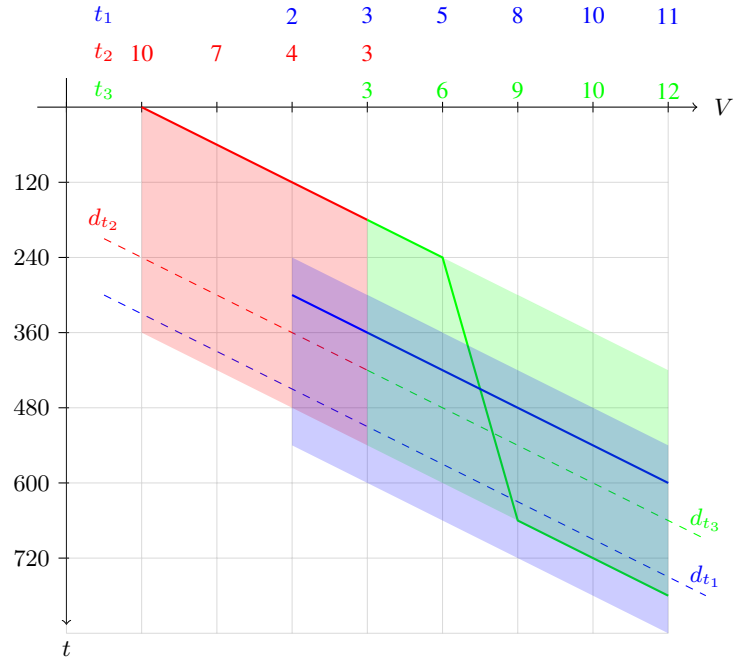
Figure~\ref{fig:times} shows the earliest and latest arrival times for nodes of the valid paths of Figure~\ref{fig:paths}.
This is indicated by the light blue, red and green areas for train lines $t_1$, $t_2$ and $t_3$, respectively.
For instance, $t_2$ may arrive at node 10 at the earliest at time point 0 and at the latest at time point 360 or $t_1$ at node 5 between 360 and 660.
Furthermore, Figure~\ref{fig:times} shows a valid schedule for train lines $t_1$, $t_2$ and $t_3$
as indicated by the blue, red and green lines, respectively.
This schedule results from the decisions that $t_2$ and $t_3$ enter $\mathit{sw}_1$ before $t_1$ and,
conversely, that $t_1$ enters $\mathit{sw}_2$ before $t_3$.
Each resource conflict is resolved by a timing constraint
indicating that the following train line enters the resource only after the preceding train line has left it plus a safety period.
In our example, this safety period, as well as the travel time for each edge, is 60 seconds.
This is reflected in the schedule since $t_2$ starts at 0 in node 10 and proceeds to node 7 at 60,
and $t_1$ only enters $\mathit{sw}_1$ at node 2 a minute after $t_3$ has left it at node 6.

In our example, there are three connections.
Our concept of connections captures transfer of passengers and cargo in various contexts,
as well as the transfer of physical trains between train lines.
The latter enables us to express cyclic train movements as can be seen with the connection of $t_2$ to $t_3$ on edges $(4,3)$ and $(3,6)$.
We call such a connection \emph{collision-free} since resource conflicts are disregarded on all connected edges that share the same resource,
and the connection furthermore requires both train lines to arrive at node 3 at the exact same time point.
This connection expresses that the same physical train is used for train line $t_2$ and $t_3$ and seamlessly transitions at node 3.\footnote{For simplicity, we assume that collision-free connections are already given with a transitive closure of all adjacent edges of the same resource and all possible other collision-free connections of other trains of that resource.}
The results are reflected in the connected paths in Figure~\ref{fig:paths} and the same arrival time at node 3 in Figure~\ref{fig:times}.
The two other connections from $t_1$ to $t_3$ on either edges $(10,11)$ and $(10,12)$ or $(10,12)$ and $(10,12)$,
on the other hand, capture a possible transfer of passengers or cargo from train line $t_1$ to $t_3$.
The connection requires $t_3$ not to arrive at 12 before $t_1$ has arrived by at least one minute at 10 so that a transfer is possible.
Since this connection does not disable resource conflicts,
this can only be achieved if $t_1$ precedes $t_2$ through $\mathit{sw}_2$ which makes $t_2$ wait between nodes 6 and 9 until $t_1$ has left.

After obtaining a valid routing and scheduling,
the solution is evaluated regarding the delay of the train lines and the quality of the paths they have taken.
The former is calculated by summing up the delay of each train line at each node in their paths.
For that purpose, a time point is defined after which the train line is considered delayed at a node.
In Figure~\ref{fig:times},
these time points are indicated by lines $d_{t_1}, d_{t_2}$ and $d_{t_3}$.
For instance, $t_3$ is delayed at nodes 9, 10 and 12,
thus accumulating a penalty of $120+120+120=360$.
For the quality of the routes, penalties are assigned to edges
and accumulated for every train line traveling that edge.
Such penalties may indicate tracks that can manage less workload, are in need of maintenance,
or are known to be a detour,
and therefore to be avoided if possible.
In our example, only edge $(1,3)$ is assigned a penalty of one,
other edges are assumed to have a penalty of zero,
therefore our routing is optimal and accumulates no routing penalties.

\subsection{Problem formalization}\label{sec:problem}

We formalize the train scheduling problem as a tuple $(N,T,C,F)$ having the following components:
\begin{itemize}
\item
$N$ stands for the railway network $(V,E,R,m,a,b)$,
where
\begin{itemize}
\item $(V,E)$ is a directed graph,
\item $R$ is a set of resources,
\item $m:E\rightarrow \mathbb{N}$ assigns the minimum travel time of an edge,
\item $a:R\rightarrow 2^E$ associates resources with edges in the railway network,
and
\item $b:R\rightarrow\mathbb{N}$ gives the time a resource is blocked after it was accessed by a train line.
\end{itemize}
\item
$T$ is a set of train lines to be scheduled on network $N$.

Each train in $T$ is represented as a tuple $(S,L,e,l,w)$,
where
\begin{itemize}
\item $(S,L)$ is an acyclic subgraph of $(V,E)$,
\item $e:S\rightarrow\mathbb{N}$ gives the earliest time a train may arrive at a node,
\item $l:S\rightarrow\mathbb{N}\cup\{\infty\}$ gives the latest time a train may arrive at a node, and
\item  $w:L\rightarrow \mathbb{N}$ is the time a train has to wait on an edge.
\end{itemize}
Note that all functions are total unless specified otherwise and we use seconds as time unit.
\item
$C$ contains connections requiring that a certain
train line $t'$ must not arrive at node $n'$ before another train line $t$ has arrived at node $n$ for at least $\alpha$ and at most $\omega$ seconds.

More precisely,
each connection in $C$ is of form $(t,(v,v'),t',(u,u'),\alpha,\omega,n,n')$
such that
$t=(S,L,e,l,w)\in T$ and $t'=(S',L',e',l',w')\in T$, $t\neq t'$,
$(v,v')\in L$, $(u,u') \in L'$,
$\{\alpha,\omega\}\subseteq \mathbb{Z}\cup\{\infty,-\infty\}$,
and either $n=v$ or $n=v'$,
as well as, either $n'=u$ or $n'=u'$.
\item
Finally, $F$ contains collision-free resource points for each connection in $C$.

We represent it as a family
\(
(F_c)_{c\in C}
\).

Each resource point in $F_c$ is of form $(t,(v,v'),t',(u,u'),r)$
and expresses that
two train lines $t$ and $t'$ are allowed to share the same resource $r$ on edges $(v,v')$ and $(u,u')$.
Connections removing collision detection are used to model splitting (or merging) of trains,
as well as reusing the whole physical train between two train lines.
More importantly, this allows us to alleviate the restriction that subgraphs for train lines are acyclic,
as we can use two train lines forming a cycle that are connected via such connections.
This can be observed in Figure~\ref{fig:paths},
where $t_2$ and $t_3$ are connected like so.

We suppose the following conditions
\begin{align*}
(t,(n,v),t',(u,u'),r) &\in F_c,\text{ if } (n,v)\in L\cap a(r), \\
(t,(v',n'),t',(u,u'),r) &\in F_c,\text{ if } (v',n')\in L\cap a(r), \\
(t,(v,v'),t',(m,u),r) &\in F_c,\text{ if } (m,u)\in L'\cap a(r),
\text{ and}\\
(t,(v,v'),t',(u',m'),r) &\in F_c,\text{ if }(u',m')\in L'\cap a(r).
\end{align*}
These assumptions ensure that adjacent edges sharing the same resource are always collision-free resource points for that connection,
because otherwise the connections could not be made if resources span over several edges.
To illustrate this,
consider two trains lines $t_a=(\{1,2\},\{(1,2)\},e,l,w)$ and $t_b=(\{2,3,4\},\{(2,3),(3,4)\},e',l',w')$ that
reuse one physical train.
All edges have a minimal travel time of six seconds and the same resource $r$ with a block time of ten seconds.
To facilitate a seamless transition from one train line to another,
a connection $(t_a,(1,2),t_b,(2,3),0,0,2,2)$ is used where both train lines blend into each other.
As both train lines ought to use the same physical train,
a collision-free point $(t_a,(1,2),t_b,(2,3),r)$ is needed to allow the connection at the given edges.
This collision-free point alone is not sufficient,
since $t_b$ would have to wait for four seconds at edge $(3,4)$ for the release of the resource $r$,
blocked by $t_a$ at edge $(1,2)$.
This is due to the fact that the blocked time of ten seconds is larger than the minimal travel time of six seconds.
Therefore, the collision-free point $((t_a,(1,2),t_b,(3,4),r))$ is necessary to
correctly execute the connection as intended,
that is, modeling one physical train among multiple train lines.

\end{itemize}

For our example in Figure~\ref{fig:paths} and Figure~\ref{fig:times}, the train scheduling problem is defined as
\begin{itemize}
\item $V=\{1,\dots,12\}$,
  \begin{itemize}
  \item $E=\{(1,3),(2,3),\dots,(10,11),(10,12)\}$,
  \item $R=\{\text{$\mathit{sw}_1$},\text{$\mathit{sw}_2$}\}\cup\{r_e\mid e\in E\}$,
  \item $m(e)=60$ and $a(r_e)=\{e\}$ for $e\in E$,
  \item $a(\mathit{sw}_1)=\{(1,3),(2,3),(4,3),(3,5),(3,6))\}$,
  \item $a(\mathit{sw}_2)=\{(8,10),(10,7),(9,10),(10,11),(10,12)\}$,
  \item $b(r)=60$ for $r\in R$,
  \end{itemize}
\item $T=\{t_1,t_2,t_3\}$ with
  \begin{itemize}
  \item $t_1=(S_1,L_1,e_1,l_1,w_1)$,
  \item $t_2=(S_2,L_2,e_2,l_2,w_2)$,
  \item $t_3=(S_3,L_3,e_3,l_3,w_3)$,
  \end{itemize}
  where $(S_1,L_1)$, $(S_2,L_2)$ and $(S_3,L_3)$ are nodes of edges and edges themselves that are solid, dotted or dashed
  in Figure~\ref{fig:paths}, respectively, and
\begin{itemize}
\item $e_1,l_1,e_2,l_2,e_3,l_3$ give the upper and lower coordinates of the colored areas in Figure~\ref{fig:times},
\item $w_1(e)=w_2(e)=w_3(e)=0$ for $e\in E$,
\end{itemize}
\item $C = \{c_1,c_2,c_3\}$
  with
  \begin{itemize}
  \item $c_1=(t_2,(4,3),t_3,(3,6),0,0,3,3)$,
  \item $c_2=(t_1,(10,12),t_3,(10,12),60,\infty,10,12)$,
  \item $c_3=(t_1,(10,11),t_3,(10,12),60,\infty,10,12)$,
  \end{itemize}
  and
\item   $F = (F_{c_1},F_{c_2},F_{c_3})$ where $F_{c_1} = \{(t_2,(4,3),t_3,(3,6),\mathit{sw}_1)\}$ and $F_{c_2}=F_{c_3}=\emptyset$.
\end{itemize}

A solution $(P,A)$ to a train scheduling problem $(N,T,C,F)$ is a pair consisting of
\begin{enumerate}
\item a function $P$ assigning to each train line the path it takes through the network,
and
\item  an assignment $A$ of arrival times to each train line at each node on their path.
\end{enumerate}
A path is a sequence of nodes, pair-wise connected by edges.
We write $v\in p$ and $(v,v')\in p$ to denote that node $v$ or edge $(v,v')$ are contained in path $p= (v_1,\dots,v_n)$,
that is, whenever $v=v_i$ for some $1\leq i\leq n$ or this and additionally $v'=v_{i+1}$, respectively.

A path $P(t)=(v_1,\dots,v_n)$ for $t=(S,L,e,l,w)\in T$ has to satisfy:
\begin{align}
  \label{cond:nodes}
  v_i \in S \text{ for }1\leq i \leq n\\
  \label{cond:path}
  (v_j,v_{j+1})\in L \text{ for }1\leq j \leq n-1\\
  \label{cond:startend}
  \mathit{in}(v_1)=0\text{ and }\mathit{out}(v_n)=0,
\end{align}
where $\mathit{in}$ and $\mathit{out}$ give the in- and out-degree of a node in graph $(S,L)$, respectively.
Intuitively, conditions \eqref{cond:nodes} and \eqref{cond:path} enforce paths to be connected and feasible for the train line in question
and Condition~\eqref{cond:startend} ensures that each path is between a possible start and end node.

An assignment $A$ is a partial function $T\times V \rightarrow \mathbb{N}$,
where $A(t,v)$ is undefined whenever $v\not\in P(t)$.
Given path function $P$, an assignment $A$ has to satisfy the conditions in~\eqref{cond:early} to~\eqref{cond:connect}:
\begin{align}
  \label{cond:early}
  A(t,v_i)&\geq e(v_i)\\
  \label{cond:late}
  A(t,v_i)&\leq l(v_i)\\
  \label{cond:travel}
  A(t,v_j)+m((v_j,v_{j+1}))+w((v_j,v_{j+1}))&\leq A(t,v_{j+1})
\end{align}
for all $t=(S,L,e,l,w)\in T$ and $P(t)=(v_1,\dots,v_n)$ such that $1\leq i \leq n, 1\leq j \leq n-1$,

either
\begin{align}
  \label{cond:conflict}
  A(t,v')+b(r)\leq A(t',u) \ \text{ or } \ A(t',u')+b(r)\leq A(t,v)
\end{align}
for all $r\in R, \{t,t'\}\subseteq T, t\ne t', (v,v')\in P(t), (u,u')\in P(t')$ with $\{(v,v'),(u,u')\}\subseteq a(r)$
whenever for all $(t,(x,x'),t',(y,y'),\alpha,\omega,n,n') \in C$ such that $(x,x')\in P(t), (y,y')\in P(t')$,
we have $(t,(v,v'),t',(u,u'),r)\not\in F_c$,

and finally
\begin{align}
  \label{cond:connect}
  \alpha\leq A(t',n')-A(t,n) \leq \omega
\end{align}
for all $(t,(v,v'),t',(u,u'),\alpha,\omega,n,n') \in C$ if $(v,v')\in P(t)$ and $(u,u')\in P(t')$.

Intuitively, conditions \eqref{cond:early}, \eqref{cond:late} and \eqref{cond:travel} ensure that a train line arrives at nodes neither too early nor too late and that waiting and traveling times are accounted for.
Furthermore, Condition~\eqref{cond:conflict} resolves conflicts between two train lines that travel edges sharing a resource,
so that one train line can only enter after another has left for a specified time span.
This condition does not have to hold if the two trains use a connection that defines a collision-free resource point for the given edges and resource.
Finally, Condition~\eqref{cond:connect} ensures that train line $t$ connects to $t'$ at node $n$ and $n'$, respectively, within a time interval from $\alpha$ to $\omega$.
Note that this is only required if both train lines use the specific edges specified in the connections.
Furthermore, note that it is feasible that $n$ and $n'$ are visited but no connection is required since one or both train lines took alternative routes.

For our solution in Figure~\ref{fig:times},
we have
\begin{itemize}
\item $P(t_1)=(2,3,5,8,10,11)$, $P(t_2)=(10,7,4,3)$, and $P(t_3)=(3,6,9,10,12)$,
and
\item  $A(t_1,2)=300,\dots, A(t_1,11)=600$,

  $A(t_2,10)=0,\dots,A(t_2,3)=180$, and

  $A(t_3,3)=180,A(t_3,6)=240, A(t_3,9)=660,\dots,A(t_3,12)=780$.
\end{itemize}

To determine the quality of a solution,
both the aggregated delay of all train lines as well as the quality of the paths through the network are taken into account.
For that purpose, we consider two functions: the delay function $d$ and route penalty function $\mathit{rp}$.
Given a train line $t=(S,L,e,l,w)\in T$ and a node $s\in S$, $d(t,s)\in \mathbb{N}$ returns the time point after which the train line $t$ is considered late at node $s$.
Note that $e(s)\leq d(t,s) \leq l(s)$.
Given an edge $e\in E$, $\mathit{rp}(e)\in \mathbb{N}$ is the penalty a solution receives for each train line traveling edge $e$.
With this, the quality of a solution $(P,A)$ is determined via the following pair:
\begin{align}
\label{eq:obj}
  \textstyle
  \big(
  \sum_{((t,v),a)\in A}{\max\{(a-d(t,v)),0\}/60},\
  \sum_{e\in \{u \mid p\in P, u\in p, e\in E\}}\mathit{rp}(e)
  \big)
\end{align}
Since delay is the more important criteria, optimization of the quality amounts to lexicographic minimization considering delay first and route penalty second.
As mentioned, our example has an accumulated delay of 360 and 0 route penalty and therefore a quality of $(360/60, 0)=(6,0)$.


\section{An ASP-based solution to real-world train scheduling}\label{sec:approach}

In this section, we present our hybrid ASP-based approach for solving the train scheduling problem.
It relies on
dedicated preprocessing techniques to reduce the problem size
as well as additional constraints and domain-specific heuristics to reduce the search space and improve solving performance.
This constitutes a significant improvement over our previous approach~\cite{abjoossctowa19a} that,
while similar in principle,
does not scale to the largest instances available.
We start by presenting said preprocessing techniques that mainly exploit redundancy in the resource distribution in the railway network.
Following that, we describe the actual hybrid encoding that makes use of this preprocessing
and furthermore reduces the amount of timing constraints by using a compressed representation of the graph.
Finally, we present optional constraints and domain-specific heuristics aiming at further improving solving performance.

\subsection{Preprocessing}

We present two techniques that reduce the complexity of the problem at hand.
While \emph{resource subsumption} detects redundant resources that can safely be removed,
\emph{resource areas} are used to simplify conflict resolution on resource conflicts.

\subsubsection{Resource subsumption}

An analysis of real-world instances revealed that resources are often contained within others, thus posing redundant constraints.
In a preprocessing step, we detect such subsumed resources and remove them from the problem specification.
Given a railway network $(V,E,R,m,a,b)$,
a resource $r_1\in R$ is subsumed by another resource $r_2\in R$,
if $a(r_1) \subseteq a(r_2)$ and $b((v,v')) \leq b((u,u'))$ for all $(v,v')\in a(r_1)$ and $(u,u')\in a(r_2)$,
and there is no collision-free resource point $(t,(v,v'),t',(u,u'),r_2) \in F_c$ for any $c\in C$.%
\footnote{If several resources are exactly the same, we keep one of them.}
Intuitively, subsumed resources are contained within another resource that induces the same or stronger timing constraints due to higher or equal blocked time and no conflict-free resource points in the overlapping area.
In our example, we can safely remove resources $\{r_{(10,7)}, r_{(10,11)}, r_{(8,10)}, r_{(9,10)}, r_{(10,12)}\}$
as each of them is subsumed by $\mathit{sw}_2$.
Note that we cannot remove any resources covered by $\mathit{sw}_1$, as it is used in the collision-free resource point
$(t_2,(4,3),t_3,(3,6),\mathit{sw}_1)$ of $c_1$.
The collision-free resource point disables conflict resolution on $\mathit{sw}_1$ for the trains $t_1$ and $t_2$,
hence, conflicts for other resources topologically contained within $\mathit{sw}_1$ are not redundant and have to be taken into account individually.

\subsubsection{Resource areas}

As seen in our example,
resources covering several edges can be exploited by identifying areas, for which
it is enough to determine the order of train lines passing through
(rather than doing pair-wise conflict resolution on edges, as done in~\cite{abjoossctowa19a}).
We call them \emph{resource areas}.
Every train line has its own set of resource areas.
It is required that every possible path through the resource area contains only edges assigned the original resource.
More precisely, a resource area $A^t_r$ over resource $r\in R$ and train line $t=(S,L,e,l,w)$ is a maximal set of edges $A^t_r \subseteq a(r)\cap L$
such that
there is no path $p=(v_1,\dots,v_n)$ in $(S,L)$ with $(u,u')\in p$ but $(u,u')\notin a(r)$
between two edges $\{(v,v_1),(v_n,v')\}\in A^t_r$.
Intuitively, this means that a resource area can only be occupied by one train at a time
independently of their chosen paths through the resource area.
Thus, other train lines using the same resource may only enter once the entire area is free.
Note that there may exist several resource areas for a resource and a train.

We define a \emph{resource coverage} $\mathcal{A}^t_r$ as a set of resource areas  over resource $r$ and train line $t$
such that $\bigcup_{A\in\mathcal{A}^t_r} A = a(r)\cap L$ and $A\cap A'=\emptyset$ for $\{A,A'\}\subseteq\mathcal{A}^t_r$.
A resource coverage $\mathcal{A}$ over sets of resources $R$ and train lines $T$ is defined as $\mathcal{A}=\bigcup_{r\in R,t\in T}\mathcal{A}^t_r$.
Intuitively, a resource coverage distributes resource areas so that every edge is covered and no edge is in two resource areas per resource and train line.

In our example, we have the resource coverage
\begin{align*}
  \mathcal{A}=\ &\{\{(v,v')\}^t_{r_{(v,v')}} \mid t\in T, r_{(v,v')} \in R \}\ \cup \\
  &\{\{(1,3),(2,3),(3,5)\}^{t_1}_{\mathit{sw}_1},\{(8,10),(10,11),(10,12)\}^{t_1}_{\mathit{sw}_2}\}\ \cup\\
  &\{\{(4,3)\}^{t_2}_{\mathit{sw}_1}\},\{(10,7)\}^{t_2}_{\mathit{sw}_2}\}\ \cup\\
  &\{\{(3,6)\}^{t_3}_{\mathit{sw}_1},\{(9,10),(10,12)\}^{t_3}_{\mathit{sw}_2}\},
\end{align*}
where $T$ and $R$ are the train lines and resources of the example in Figure~\ref{fig:paths}, respectively.

We have three non singleton sets in the coverage.
Given the train lines $t_1$, $t_3$ and resource $\mathit{sw}_2$ from our example,
we have to decide whether $t_1$ enters $\{(8,10),(10,11),(10,12)\}^{t_1}_{\mathit{sw}_2}$ first or $t_3$ enters $\{(9,10),(10,12)\}^{t_3}_{\mathit{sw}_2}$ first.
Without the use of resource areas, we would need to make a decision for every element in the cross product $\{(8,10),(10,11),(10,12)\}^{t_1}_{\mathit{sw}_2} \times \{(9,10),(10,12)\}^{t_3}_{\mathit{sw}_2}$,
leaving us with 6 decisions instead of one.
Note that there is no unique resource coverage for every graph, but different maximal subsets could be chosen.
\begin{figure}
\begin{algorithm}[H]
\SetKwInOut{Input}{Input}\SetKwInOut{Output}{Output}
\Input{Resource $r$ and train line $t=(S,L,e,l,w)$.}
\Output{A resource coverage $A^t_r$.}
$A^t_r\leftarrow\emptyset$\;
\While{$\bigcup_{A\in A^t_r}{A} \neq a(r)\cap S$\label{algo:start}}
{
  $A\leftarrow \emptyset$\label{algo:empty}\;
  \ForEach{$(v,v') \in (a(r)\cap S) \setminus \bigcup_{A\in A^t_r}{A}$\label{algo:inc}}
  {
    \uIf{$\textit{isRA}(A\cup\{(v,v')\},r,S,L)$\label{algo:isRA}}
    {
      $A\leftarrow A \cup \{(v,v')\}$\label{algo:add}\;
    }
  }
}
\end{algorithm}
  \caption{Algorithm to compute resource coverage.\label{fig:algo}}
\end{figure}

To determine resource areas, we use the greedy algorithm in Figure~\ref{fig:algo} for every resource $r$ and train line $t=(S,L,e,l,w)$.
We incrementally create a resource area $A$ by adding yet unused edges $(v,v')$ to it in Line~\ref{algo:inc}.
The function $\textit{isRA}(A,r,S,L)$ in Line~\ref{algo:isRA} checks that no path in $(S,L)$ between edges in $A$ contains edges not assigned $r$.
Intuitively, this means that once a train line entered a resource area, it is not able to leave it and reenter it again.
If this is the case, $(v,v')$ is added to $A$ in Line~\ref{algo:add}.
This is repeated until we cannot add further edges to the current resource area and
then restart the procedure creating a new area in lines~\ref{algo:start} and~\ref{algo:empty} until we achieve full resource coverage.

We outline how we use resource areas to derive solutions fulfilling Condition~\eqref{cond:conflict} in Section~\ref{sec:encoding}.

\subsection{Fact format}
\label{sec:fact_format}

A train scheduling problem $(N,T,C,F)$ with $N=(V,E,R,m,a,b)$ is represented by the facts
\begin{lstlisting}[mathescape,numbers=none]
   tl($t$).    edge($t$,$v$,$v'$).    m($(v,v')$,$m((v,v'))$).    w($t$,$(v,v')$,$m((v,v'))$).
\end{lstlisting}
for each $t=(S,L,e,l,w)\in T$ and $(v,v')\in L$.

For every $s\in S$, we add
\begin{lstlisting}[mathescape,numbers=none]
   e($t$,$s$,$e(s)$).     l($t$,$s$,$l(s)$).
\end{lstlisting}
Additionally, either
\begin{lstlisting}[mathescape,numbers=none]
   start($t$,$s$).     |\textnormal{or}|     end($t$,$s$).
\end{lstlisting}
is added, if $\mathit{in}(s)=0$ or $\mathit{out}(s)=0$ in $(S,L)$, respectively.
We assign unique terms to each train line for identifiability.

For example, the facts
\begin{lstlisting}[numbers=none]
 tl(t1).         edge(t1,1,3).    m((1,3),60).    w(t1,(1,3),0).
 e(t1,1,240).    l(t1,1,540).
 start(t1,1).
\end{lstlisting}
express that train line \texttt{t1} may travel between nodes \texttt{1} and \texttt{3} taking at least 60 seconds, waiting on this edge for 0 seconds,
and arrives between time points 240 and 540 at node \texttt{1}, which is a possible start node.

Furthermore, we add
\begin{lstlisting}[mathescape,numbers=none]
   resource($r$,($v$,$v'$)).     b($r$,$b(r)$).
\end{lstlisting}
for $r\in R$ and $(v,v')\in a(r)$.
Akin to train lines, resources are assigned unique terms to distinguish them.

For example, facts
\begin{lstlisting}[mathescape,numbers=none]
   resource(sw1,(1,3)).    resource(sw1,(4,3)).    b(sw1,60).
\end{lstlisting}
assign resource \texttt{sw1} to edges $(1,3)$ and $(4,3)$ and the resource is blocked for 60 seconds after a train line has left it.

Finally, we add
\begin{lstlisting}[mathescape,numbers=none]
   connection($c$,$t$,($v$,$v'$),$t'$,($u$,$u'$),$\alpha$,$\omega$,$n$,$n'$).
\end{lstlisting}
for all $(t,(v,v'),t',(u,u'),\alpha,\omega,n,n')\in C$,
where $c$ acts as an identifier,
and
\begin{lstlisting}[mathescape,numbers=none]
   free($c$,$t$,($v$,$v'$),$t'$,($u$,$u'$),$r$).
\end{lstlisting}
for all $(t,(v,v'),t',(u,u'),r)\in F_c$.

For instance, the transfer of the physical train from train line \texttt{t2} to \texttt{t3} at node $3$ is encoded by
\lstinline{connection(1,t2,(4,3),t3,(3,6),0,0,3,3).}
This requires both train lines to be at node \code{3} at the exact same time.
Thus, we have to additionally provide the fact
\lstinline{free(1,t2,(4,3),t3,(3,6),sw1)}
to allow for a shared use of the resource, so that the train lines can connect.
This can be done safely since in reality only one train exists for both train lines,
which rules out collisions.

For a resource area $A\in\mathcal{A}^t_r$, we generate facts
\begin{lstlisting}[mathescape,numbers=none]
   ra($t$,$r$,$a$,($v$,$v'$)).
\end{lstlisting}
for $(v,v')\in A$, resource $r \in R$ and train line $t=(S,L,e,l,w)\in T$.
We assign a unique term $a$ to distinguish resource areas for the same resource and train line.
Furthermore, for every resource area $A\in\mathcal{A}^t_r$, we add
\begin{lstlisting}[mathescape,numbers=none]
   e_ra($t$,$r$,$a$,$e$).     l_ra($t$,$r$,$a$,$l$).
\end{lstlisting}
where $e=min\{e(v))\mid (v,v')\in A\}$ and $l=max\{l(v')\mid (v,v')\in A\}$ to represent the earliest entry and latest exit times for that resource area.
In our example, the earliest entry time for resource area $\{(1,3),(2,3),(3,5)\}^{t_1}_{\mathit{sw}_1}$ is 0, while the latest exit time is 660.

Given delay and route penalty functions $d$ and $\mathit{rp}$, we add
\begin{lstlisting}[mathescape,numbers=none]
   potlate($t$,$s$,$u$,$p$).     penalty($m$,$\mathit{rp}(m)$).
\end{lstlisting}
for $t=(S,L,e,l,w)\in T,s\in S,m\in L$ with $\{u,p\}\subseteq \mathbb{N}, d(t,s) < u \leq l(t,s)$ to evaluate a solution.

The collection of facts for our example instance can be found in~\ref{appendix:facts} in Listing~\ref{enc:facts}.

\subsection{Encoding}
\label{sec:encoding}

In the following, we describe the general problem encoding.
We separate it into three parts handling path constraints, conflict resolution and scheduling.

\begin{minipage}{\linewidth}
\begin{lstlisting}[label=lst:routing,caption={Encoding of path constraints.},firstnumber= 1]
1 { visit(T,V)      : start(T,V)   } 1 :- tl(T).|\label{enc:start}|
1 { route(T,(V,V')) : edge(T,V,V') } 1 :- visit(T,V),
                                          not end(T,V).|\label{enc:route}|
visit(T,V') :- route(T,(V,V')).|\label{enc:visit}|
\end{lstlisting}
\end{minipage}
The first part of the encoding in Listing~\ref{lst:routing} covers routing.
First, exactly one valid start node is chosen for each train line to be visited (Line~\ref{enc:start}).
From a node that is visited by a train line and is not an end node,
an edge in the relevant subgraph is chosen as the next route (Line~\ref{enc:route}).
The new route in turn leads to a node being visited by the train line (Line~\ref{enc:visit}).
This way, each train line is recursively assigned a valid path.
Since those paths begin at a start node, finish at an end node and are connected via edges valid for the respective train lines,
conditions \eqref{cond:path} and \eqref{cond:startend} are ensured.

\begin{minipage}{\linewidth}
\begin{lstlisting}[label=lst:conflicts,caption={Encoding of conflict resolution.},firstnumber=5]
enter_ra(T,R,A,V) :-|\label{enc:enter_ra}|
   ra(T,R,A,(V,V')), route(T,(V,V')),
   1 #sum {1 : edge(T,V'',V), not ra(T,R,A,(V'',V));
           1 : start(T,V)},
   not route(T,(V'',V)) : ra(T,R,A,(V'',V)).
leave_ra(T,R,A,V') :-
   ra(T,R,A,(V,V')), route(T,(V,V')),
   1 #sum {1 : edge(T,V',V''), not ra(T,R,A,(V',V''));
           1 : end(T,V')},
   not route(T,(V',V'')) : ra(T,R,A,(V',V'')).|\label{enc:leave_ra}|

free_a(T,T',R,A,A') :- free(I,T,(V,V'),T',(U,U'),R),|\label{enc:start_free_a}|
                       ra(T,R,A,(V,V')), ra(T',R,A',(U,U')),
                       connection(I,T,(X,X'),T',(Y,Y'),_,_,_,_),
                       route(T,(X,X')), route(T',(Y,Y')).|\label{enc:end_free_a}|

shared(T,T',R,A,A') :- e_ra(T,R,A,E), e_ra(T',R,A',E'), |\label{enc:start_shared}|
                       not free_a(T,T',R,A,A'),|\label{enc:nonfree}|
                       l_ra(T,R,A,L), b(R,B), T != T',
                       E <= E', E' < L+B.
shared(T',T,R,A',A)  :- shared(T,T',R,A,A').|\label{enc:end_shared}|

{ seq(T,T',R,A,A') } :-  shared(T,T',R,A,A'), T < T'.|\label{enc:startseq}|
  seq(T',T,R,A',A)   :-  shared(T,T',R,A,A'),
                         not seq(T,T',R,A,A').|\label{enc:endseq}|

\end{lstlisting}
\end{minipage}
The next part of the encoding shown in Listing~\ref{lst:conflicts} detects and resolves resource conflicts.
Basically, a resource conflict exists, if two train lines each have an edge in their respective subgraph that is assigned the same resource,
and time intervals overlap in which the train lines enter and leave the edges in question,
extended by the time the resource is blocked.
We improve upon this edge-centric conflict resolution via resource areas.
In lines~\ref{enc:enter_ra}--\ref{enc:leave_ra}, we first compute entry and exit nodes leading in and out of a resource area (depending on the chosen route).
A conflict is possible, if two train lines each have a resource area in their respective subgraph that is assigned the same resource,
and the time intervals overlap in which the train lines may enter and leave the areas in question,
extended by the time the resource is blocked (lines~\ref{enc:start_shared}--\ref{enc:end_shared}).
We resolve the conflict by making a choice which train line passes through this area first (lines~\ref{enc:startseq} and~\ref{enc:endseq}).

As a collision-free resource point $(t,(v,v'),t',(u,u'),r)$ does not trigger a resource conflict,
we prevent the involved resource areas $A^t_r$ and $A^{t'}_r$ with $(v,v')\in A^t_r$ and $(u,u')\in A^{t'}_r$ from creating conflicts by
deriving collision-free resource areas in lines~\ref{enc:start_free_a}--\ref{enc:end_free_a}.
It is safe to do so,
as it is necessary for a collision-free connection that all edges adjacent to $(v,v')$ or $(u,u')$ with the same resource are also collision-free resource points (recall Section~\ref{sec:problem}).
We exclude these areas from the conflict resolution in Line~\ref{enc:nonfree}.

Note that we require much less decisions for resolving resource conflicts by using resource areas.
It is possible that several or even all edges in two resource areas with the same resource induce a conflict.
Naively, a decision would have to be made for all possible combination of those edges.
Instead of having this quadratic blowup, we only have to resolve the resource conflict between entire resource areas.
For our example,
this reduces the number of decisions necessary to resolve all possible conflicts from 15 to 5.

To represent the arrival time of a train line $t=(S,L,e,l,w)$ at node $v\in S$,
we project the graph $(S,L)$ to a possibly smaller version of it using a topological ordering.
In essence, instead of assigning an arrival time to each possible node a train line could visit,
we assign the arrival time to the progress the train line has made relative to its subgraph.
For that purpose, we utilize the \emph{height} of a node that indicates the maximum number of edges the train line has to travel to reach this node,
(or, in other words,
in an acyclic graph, the height of a node is the length of the longest path from any root node to this node).
Obviously, several nodes can have the same height whenever they are on parallel paths through the subgraph.
Formally, the height of a node $v\in S$ for train line $t$ is defined as follows:
\[
  h^t(v)=
  \begin{cases*}
    0                                      & if $\mathit{in}(v)=0$\\
    \max\{h^t(v')) \mid (v',v) \in L\} + 1 & otherwise
  \end{cases*}
\]
The arrival times are now represented using integer variables designated by pairs of form \code{($t$,$h^t(v)$)}
indicating arrival time of train line $t$ at height $h^t(v)$\footnote{Cf.~Section~\ref{sec:background} on using pairs as identifiers.}
This reduces the number of variables and difference constraints needed whenever trains are routed over nodes with the same height.
For the running example, nodes 1 and 2 from train line $t_1$ now collapse to one variable $(t_1,0)$,
as only one of the two nodes can be used in a final routing of the train line.
Instead of introducing arrival times for node 1 and 2,
we only introduce an arrival time $(t_1,0)$ where $0=h^{t_1}(1)=h^{t_1}(2)$, independent of the chosen path.
The number of integer variables for the arrival times of $t_1$ is therefore reduced from 8 to 6 in contrast to using node names such as \code{($t$,$v$)}.

Note that we use ground terms in the remainder of this paper to describe our rules, while their actual encoding uses variables.
Also,
we sometimes mix mathematical notation for train lines and nodes (\textit{italic})
and their identifying terms in the encoding (\texttt{typewriter}) and use them interchangeably.
\pagebreak
\begin{lstlisting}[label=lst:scheduling,caption={Encoding of scheduling.},firstnumber=30]
h(T,V,0)   :- start(T,V).|\label{enc:start_height}|
h(T,V,M+1) :- edge(T,_,V),
              M = #max {N : h(T,V',N), edge(T,V',V)}, M != #inf.|\label{enc:end_height}|

min_e(T,N,M) :- h(T,_,N),|\label{enc:min_arrive_start}|
                M != #sup, M = #min {E : h(T,V,N), e(T,V,E)}.|\label{enc:min_arrive_end}|
max_l(T,N,M) :- h(T,_,N),|\label{enc:max_arrive_start}|
                M != #inf, M = #max {L : h(T,V,N), l(T,V,L)}.|\label{enc:max_arrive_end}|
&diff{ 0-(T,N) } <= -E :- min_e(T,N,E).|\label{enc:post_min}|
&diff{ (T,N)-0 } <=  L :- max_l(T,N,L).|\label{enc:post_max}|
&diff{ 0-(T,N) } <= -E :- e(T,V,E), visit(T,V), h(T,V,N),
                          min_e(T,N,M), M<E.|\label{enc:post_min_dyn}|
&diff{ (T,N)-0 } <=  L :- l(T,V,L), visit(T,V), h(T,V,N),
                          max_l(T,N,M), M>L.|\label{enc:post_max_dyn}|

time(T,(V,V'),(N,N'),D) :- edge(T,V,V'), h(T,V,N), h(T,V',N'),|\label{enc:start_traveltime}|
                           D=#sum{ M,m : m((V,V'),M);
			           W,w : w(T,(V,V'),W) }.
min_time(T,(N,N'),M) :-|\label{enc:start_nodebridge}|
   edge(T,V,V'), h(T,V,N), h(T,V',N'),
   not edge(T,U,U') : h(T,U,N), h(T,U',O), O != N';
   not edge(T,U,U') : h(T,U,O), h(T,U',N'), O != N;|\label{enc:end_nodebridge}|
   M = #min {D : time(T,_,(N,N'),D)}.|\label{enc:end_traveltime}|
&diff{ (T,N)-(T,N') } <= -D :- min_time(T,(N,N'),D).|\label{enc:derive_mintravel}|
&diff{ (T,N)-(T,N') } <= -D :- |\label{enc:start_derive_travel}|
   route(T,(V,V')), h(T,V,N), h(T,V',N'),
   time(T,(V,V'),(N,N'),D),
   D > #max {#inf; M : min_time(T,(N,N'),M)}.|\label{enc:end_derive_travel}|

&diff{ (T,N)-(T',N') } <= -B :- |\label{enc:start_seqdiff}|
   seq(T,T',R,A,A'), shared(T,T',R,A,A'),
   h(T,V,N), h(T',U,N'),
   leave_ra(T,R,A,V), enter_ra(T',R,A',U), b(R,B).|\label{enc:end_seqdiff}|

&diff{ (T,N)-(T',N') } <= -M :-|\label{enc:start_conndiff}|
   connection(I,T,(V,V'),T',(U,U'),M,_,X,Y),
   h(T,X,N), h(T',Y,N'), route(T,(V,V')), route(T',(U,U')).|\label{enc:route_conndiffmin}|
&diff{ (T',N')-(T,N) } <= M  :-
   connection(I,T,(V,V'),T',(U,U'),_,M,X,Y), M != #inf,
   h(T,X,N), h(T',Y,N'), route(T,(V,V')), route(T',(U,U')).|\label{enc:route_conndiffmax}|

\end{lstlisting}

Listing~\ref{lst:scheduling} displays the encoding of scheduling via difference constraints.
An atom \lstinline[mathescape]{h($t$,$v$,$n$)} assigns a node $v$ its height $n=h^{t}(v)$ for train line $t$ and
is computed in lines~\ref{enc:start_height} and~\ref{enc:end_height}.
Note that we use pairs $(t,h)$ as names for integer variables whose values indicate the arrival time of train $t$ at height $h$.
The next part of the encoding ensures that earliest and latest arrival times, viz. $e(v)$ and $l(v)$,
for a train line $t=(S,L,e,l,w)$ on a node $v\in S$ are respected
by setting lower and upper bounds for the integer variable \code{($t$,$h^t(v)$)} (lines 40-43).
Even though, the lower and upper bound of variable \code{($t$,$h$)} depend on the node that is actually visited
as there possibly exist several nodes $v\in S$ on alternative paths with the same height $h$ (the same maximum path length from a root node),
we can take advantage of the height-based naming scheme by setting the lower and upper bound of variable \code{($t$,$h$)} to $min\{e(v)\mid v\in S, h^t(v)=h \}$ and $max\{l(v)\mid v\in S, h^t(v)=h\}$, respectively,
independently of the chosen route.
For all nodes with a greater time for the earliest arrival (lesser time for the latest arrival),
an additional constraint is derived depending on whether such a node is visited.
This restricts the search space in two ways.
First, constraints for earliest and latest arrival times are independent of routing for nodes with a minimal earliest or maximal latest arrival time,
second, regardless of routing, a constraint is added constituting the lower bound of the earliest (upper bound for the latest)
arrival time at a certain height.
Lines~\ref{enc:min_arrive_start} to~\ref{enc:max_arrive_end} compute said minimal and maximal arrival times for a train line at a certain height.
Difference constraints representing these upper and lower bounds are derived in lines~\ref{enc:post_min} and~\ref{enc:post_max}
and are independent of the chosen route.
More precisely, for train line $t$ at height $h$, we derive
\begin{lstlisting}[mathescape,numbers=none]
   &diff{0-$(t,h)$} <= $-e$ |\textnormal{ and }| &diff{0-$(t,h)$}<= $-l$ |\textnormal{,}|
\end{lstlisting}
where $e=\min\{e(v)\mid v\in S, h^t(v)=h \}$ and $l=\max\{l(v)\mid v\in S, h^t(v)=h\}$.
For nodes $v$, where $e(v)=e$ and $l(v)=l$, we therefore ensure $e(v) \leq (t,h^t(v)) \leq l(v)$ and in turn condition \eqref{cond:early} and \eqref{cond:late}.
In case train line $t$ visits a node $v$ for which either $e(v)>e$ or $l(v)<l$,
we have to additionally derive
\begin{lstlisting}[mathescape,numbers=none]
   &diff{0-$(t,h^t(v))$} <= $-e(v)$ |\textnormal{ or }| &diff{$(t,h^t(v))$-0} <= $l(v)$|\textnormal{\,, respectively.}|
\end{lstlisting}
This again ensures $e(v) \leq (t,h^t(v)) \leq l(v)$ and in turn condition \eqref{cond:early} and \eqref{cond:late} to hold.

The next part of the encoding guarantees the travel and waiting time between two nodes $(v,v')\in L$ for a train line $t=(S,L,e,l,w)$.
We call adjacent heights $(n,n+1)=(h^t(v),h^t(v'))$ \emph{exclusively connected},
if there is no edge $(u,u')\in L$ such that $h^t(u)=n$ and $h^t(u')\neq n+1$ or $h^t(u')=n+1$ and $h^t(u)\neq n$.
For these heights $(n,n+1)$ the minimum of the travel plus waiting time can be set independently of the chosen route.
An exclusively connected pair $(n,n+1)$ represents a set of edges $\{(v_1,v_2),\dots,(v_{m-1},v_m)\}\subseteq L$ of which at most one edge can be visited by a train line.
The minimum of travel plus waiting time of these edges can therefore be guaranteed independently of routing,
as it constitutes a lower bound.
For any edge $(v,v')\in L$, where either the travel plus waiting time is greater than the minimum across all edges for exclusively connected heights
$(h^t(v),h^t(v'))$, or $(h^t(v),h^t(v'))$ is not exclusively connected,
the constraint for minimal distance between $(t,v)$ and $(t,v')$ depends on the routing.
In the worst case, no exclusively connected heights exist and all constraints are dependent on the routing.
In our running example, edges $(1,3)$ and $(2,3)$ for train line $t_1$ result in the same pair of exclusively connected heights $(0,1)$.
This means that the condition $(t_1,0) + 60 \leq (t_1,1)$ holds independently of routing, as 60 is the minimum of travel plus waiting time for these edges.
If edge $(2,3)$ had a waiting time $w((2,3))=10$, the condition $(t_1,0) + 70 \leq (t_1,1)$ needs to hold additionally,
if edge $(2,3)$ is visited.
Note that the route independent constraint only poses a lower bound on the relation of the two arrival times and is therefore subsumed by this new constraint.
Given a train line $t=(S,L,e,l,w)$, lines~\ref{enc:start_traveltime}--\ref{enc:end_traveltime},
first, compute the travel plus waiting time $\mathit{time}^t_{(v,v')}=m((v,v'))+w((v,v'))$ for all edges $(v,v')\in L$,
secondly, the minimum of travel plus waiting time $\mathit{min\_time}^t_{(n,n+1)}=\min\{\mathit{time}^t_{(v,v')} \mid (v,v')\in L, h^t(v)=n, h^t(v')=n+1 \}$ for all exclusively connected heights $(n,n+1)$,
and finally, the difference constraint atom
\begin{lstlisting}[mathescape,numbers=none]
   &diff{$(t,n)$-$(t,n+1)$}<= $-min\_time^t_{(n,n+1)}$
\end{lstlisting}
is derived in Line~\ref{enc:derive_mintravel}.
For any edge $(v,v')\in L$, where either $time^t_{(v,v')}>min\_time^t_{(h^t(v),h^t(v'))}$ holds, or $(h^t(v),h^t(v'))$ are not exclusively connected,
the difference constraint atom
\begin{lstlisting}[mathescape,numbers=none]
   &diff{$(t,h^t(v))$-$(t,h^t(v'))$} <= $-time^t_{(v,v')}$
\end{lstlisting}
is derived, whenever the train line is actually routed over edge $(v,v')$ (lines~\ref{enc:start_derive_travel}--\ref{enc:end_derive_travel}).
For all exclusively connected heights $(n,n+1)$,
the condition $(t,n) + \mathit{min\_time}^t_{(n,n+1)} \leq (t,n+1)$ is provided.
This satisfies Condition~\eqref{cond:travel} for all edges $(v,v')\in L$, where $h^t(v)=n$, $h^t(v')=n+1$ and
$\mathit{min\_time}^t_{(n,n+1)}=m((v,v')) + w((v,v'))$.
As the minimum time is a lower bound, no constraints are violated if these edges are not visited.
For all visited edges $(v,v')\in L$ where $h^t(v)=n$, $h^t(v')=n+1$ and $\mathit{min\_time}^t_{(n,n+1)}<m((v,v')) + w((v,v'))$ or $(h^t(v),h^t(v'))$ is not exclusively connected,
condition $(t,n) + m(v,v') + w((v,v')) \leq (t,n+1)$ for all visited edges $(v,v')$ has to hold.
This ensures Condition~\eqref{cond:travel}.
As for the earliest and latest arrival times above, by using a formulation based on heights,
we may either gain a lower bound or provide timing constraints independently of the routing,
thus restricting the search space.

The rule in lines \ref{enc:start_seqdiff}--\ref{enc:end_seqdiff} in Listing~\ref{lst:scheduling}
utilizes conflict detection and resolution from Listing~\ref{lst:conflicts}.
We derive the difference constraint atom
\begin{lstlisting}[mathescape,numbers=none]
   &diff{$(t,h^t(v))$-$(t',h^{t'}(u))$} <= $-b(r)$
\end{lstlisting}
expressing that $t$ leaves the resource area $A^t_r$ using a node $v$ before $t'$ enters $A^{t'}_r$ via a node $u$,
given the blocked time $b(r)$ of a resource $r$
and the decision that train line $t$ on resource area $A^t_r$ takes precedence over $t'$ on resource area $A^{t'}_r$.
As all edges in these resource areas have the same resource $r$,
and once inside a resource area a train cannot leave the resource area again,
the order of passing all edges is the same.
Therefore, this constraint supersedes all constraints $(t,h^t(v))+b(r) \leq (t,h^{t'}(u'))$ for all edges $(v,v')\in A^t_r$, $(u,u')\in A^{t'}_r$,
and captures Condition~\eqref{cond:conflict}.

Finally, lines~\ref{enc:start_conndiff}--\ref{enc:route_conndiffmax} handle connections $(c,t,(v,v'),t',(u,u'),\alpha,\omega,v'',u'') \in C$.
Difference constraint atoms
\begin{lstlisting}[mathescape,numbers=none]
   &diff{$(t,h^t(v''))$-$(t',h^{t'}(u''))$} <= $-\alpha$ |\textnormal{ and }|
   &diff{$(t',h^{t'}(u''))$-$(t,h^t(v''))$} <= $\omega$
\end{lstlisting}
are derived (lines~\ref{enc:route_conndiffmin} and~\ref{enc:route_conndiffmax})
guaranteeing $\alpha \leq (t',h^{t'}(u'')) - (t,h^t(v'')) \leq \omega$,
and thus Condition~\eqref{cond:connect},
if $(v,v')$ is visited by train line $t$ and $(u,u')$ is visited by $t'$.

\subsection{Optimization}
\label{sec:opt}

As mentioned above, we use instances of \code{potlate}/4 to indicate when a train line is considered late at a node and how to penalize its delay.
For this purpose, we choose sets $D_{t,v}\subseteq\mathbb{N}$ whose elements act as thresholds for arrival time of train line $t$ at node $v$.
Given delay function $d$,
$d(t,v) \leq u \leq l(v)$ for every $u\in D_{t,v}$, train line $t=(S,L,e,l,w)\in T$ and $v\in S$.
We create facts $\code{potlate($t$,$v$,$u$,$u-u'$)}$
      for $u,u'\in D_{t,v}$ with $u'<u$ such that there is no $u''\in D_{t,v}$ with $u' < u'' <u$.
We add {\code{potlate($t$,$v$,$u$,$u-d(t,v)$)}} for  $u=\min(D_{t,v})$.
Intuitively, we choose the penalty of a potential delay as the difference to the previous potential delay,
or, if there is no smaller threshold, the difference to the time point after which the train is considered delayed.
This way, the sum of penalties amounts to a lower bound on the train line's actual delay in seconds.
For example, for $D_{t,v}=\{6,10,14\}$ and $d(t,v)=5$,
we create facts {\code{potlate($t$,$v$,6,1)}}, {\code{potlate($t$,$v$,10,4)}} and {\code{potlate($t$,$v$,14,4)}}.
Now, if $t$ arrives at $v$ at 12, it is above thresholds 6 and 10 and should receive a penalty of 5.
This penalty is a lower bound on the actual delay of 7,
and we know that the value has to be between 5 and 9 since the next threshold adds a penalty of 4.
This method approximates the exact objective function in~\eqref{eq:obj} in two ways.
First, we do not divide by 60 and penalize in minutes since this would lead to rounding problems.
Second, our penalty only gives a lower bound to the actual delay if thresholds are more than one second apart.
While our method allows us to be arbitrarily precise in theory,
in practice, creating a threshold for each possible second of delay leads to an explosion in size.
We employ two schemes for generating sets $D_{t,v}$ given $t=(S,L,e,l,w)\in T$, $v\in S$ and delay function $d$.

\subsubsection{Binary approximation} Binary approximation detects if a train is a second late and penalizes it by one,
therefore, only the occurrence of a delay is detected while its amount is disregarded.

We set $D_{t,v}=\bind{t}{v}=\{d(t,v)+1\}$.

\subsubsection{Linear approximation}
Linear approximation evenly distributes thresholds $m$ seconds apart across the time span in which a delay might occur.
Here, if train line $t$ arrives at or after $n*m + d(t,v)$ at $v$,
we know that the real delay is between $n*m$ and $(n+1)*m$ for $n\in\mathbb{N}\setminus \{0\}$.
We also add \bind{t}{v} to detect solutions without delay.

Accordingly,
we set $D_{t,v} = \bind{t}{v} \cup \lind{t}{v}{m}$ with $\lind{t}{v}{m} = \{y\in\mathbb{N} \mid y=x*m + d(t,v), x\in\mathbb{N}\setminus\{0\}, y \leq l(v) \}$.

\subsubsection{Encoding}
\begin{minipage}{\linewidth}
\begin{lstlisting}[label=lst:min,caption={Delay and routing penalty minimization.},firstnumber=70,firstline=5]
{ late(T,V,D,W) : visit(T,V) } :- potlate(T,V,D,W).|\label{enc:min_late}|

next(T,V,D,D') :- potlate(T,V,D,_), potlate(T,V,D',_), D<D',|\label{enc:min_startchain}|
                  not potlate(T,V,D'',_) : potlate(T,V,D'',_),
                  D''>D, D''<D'.
:- not late(T,V,D,_), late(T,V,D',_), next(T,V,D,D').|\label{enc:min_endchain}|

topo_late(T,N,D,W) :- late(T,V,D,W), h(T,V,N).|\label{enc:min_topo}|

&diff{ 0-(T,N) } <= -D :- topo_late(T,N,D,W).|\label{enc:min_startdiff}|
&diff{ (T,M)-0 } <=  N :- not topo_late(T,M,D,W),
                          potlate(T,V,D,W),
                          N=D-1, visit(T,V), h(T,V,M).|\label{enc:min_enddiff}|

#minimize{ W@1,T,N,D : topo_late(T,N,D,W) }.|\label{enc:minimize_delay}|
#minimize{ P@0,T,E   : penalty(E,P), route(T,E) }.|\label{enc:minimize_penalty}|
\end{lstlisting}
\end{minipage}
Given thresholds $D_{t,v}$ for all train lines and nodes and the corresponding instances of predicate \code{potlate}/4,
Listing~\ref{lst:min} shows the implementation of the delay minimization.
The basic idea is to use regular atoms to choose whether a train line is delayed on its path for every potential delay (Line~\ref{enc:min_late} in Listing~\ref{lst:min}).
In lines~\ref{enc:min_startchain}--\ref{enc:min_endchain}, we order the given thresholds for every train line and node.
By enforcing downwards adjacent thresholds to hold as well
(being late for 4 minutes implies being late for 3 minutes which in turn implies being late for 2 minutes etc.),
we immediately exclude solutions where this semantic condition fails, thus improving propagation.
Originally, the semantics of the \code{late} atoms (being late for $x$ minutes) is only given via the difference constraint atom introduced later.
By directly encoding the implicit ordering of natural numbers using regular ASP atoms, we leverage the efficient Boolean constraint solving techniques of the ASP solver.
Line~\ref{enc:min_topo} adjusts the \code{late} atoms to the topological ordering.
In our example, this means that train line $t_1$ being late 1 minute on node 11 uses the same variable as being late 1 minute on node 12 thus reducing the amount of variables that are needed to capture delay.
Finally, we derive difference constraint atoms expressing this information (lines~\ref{enc:min_startdiff}--\ref{enc:min_enddiff}),
and ultimately using the regular atoms in a standard minimize statement (Line~\ref{enc:minimize_delay}).
In detail, for every atom {\code{potlate($t$,$v$,$u$,$w$)}}, an atom {\code{late($t$,$v$,$u$,$w$)}} can be chosen if $t$ visits $v$.
For every {\code{late($t$,$v$,$u$,$w$)}} we derive {\code{topo\_late($t$,$\mathit{h}_t(v)$,$u$,$w$)}}.
If {\code{topo\_late($t$,$\mathit{h}_t(n)$,$u$,$w$)}} is chosen to be true,
a difference constraint atom {\code{\&diff\{0-$(t,\mathit{h}_t(v))$\}<=} $-u$} is derived expressing $(t,\mathit{h}_t(v))\geq u$ and, therefore, that $t$ is delayed at the visited node $v$ at threshold $u$.
Otherwise, {\code{\&diff\{$(t,\mathit{h}_t(v))$-0\}<=} $u-1$} becomes true so that $\code{$(t,\mathit{h}_t(v))$} < u$ holds.
The difference constraint atoms ensure that if the truth value of a \code{late} atom is decided, the schedule has to reflect this information.
The minimize statement then sums up and minimizes the penalties of the \code{late} atoms that are true.

Finally, Line~\ref{enc:minimize_penalty} shows the straightforward encoding of the routing penalty minimization.
The minimize statement merely collects the paths of the train lines, sums up their penalties, and minimizes this sum.
This is done on a lower level than delay minimization leading to lexicographic optimization minimizing delay first and route penalty second.

\subsection{Optional Constraints and Heuristics}
\label{sec:optcons}

While the above encoding can be used as is,
additional constraints can be added to further restrict the search space in the hope of improving solving performance.
To this end,
we rely on the fact that~\clingoDL~consists of a Boolean search engine (\clingo) and a dedicated difference constraints propagator.
While some atoms are purely Boolean, others have a semantics in terms of difference constraints.
In the following, we transfer the knowledge represented in these difference constraints (such as transitivity and acyclicity) back to the logic program,
thereby improving the search in the ASP part of the problem.
Finally, a domain-specific heuristic is represented that prefers sequences of train lines reducing the likelihood of conflicts.

\subsubsection{Resource overlap I}

Resource areas with different resources overlap if both contain at least one common edge.
If two train lines are each routed over such an overlap, the sequence of entering the respective pairs of resource areas has to be the same
since it is not possible for trains to overtake each other throughout both overlapping resource areas.
We exploit this by detecting such overlaps and using them to add constraints that remove solution candidates for which the sequences do not match,
thus reducing the search space.
More precisely, given two resources $r$ and $r'$, train lines $t$ and $t'$,
and resource areas $A\in \mathcal{A}^t_r$, $A' \in \mathcal{A}^t_{r'}$, $B\in \mathcal{A}^{t'}_r$, and $B'\in \mathcal{A}^{t'}_{r'}$,
if train line $t$ leaves resource area $A$ before train $t'$ enters its resource area $B$, $t$ visits an edge $(v,v')\in A\cap A'$
and $t'$ visits and edge from $(u,u')\in B\cap B'$ (i.e.\ both trains use the overlapping areas),
then train $t$ also has to leave resource area $A'$ before train $t'$ enters resource area $B'$.

As an example, we use a straight line of three tracks, where nodes are ordered adjacently from left to right.
Consider train lines $t_a = (S,L,e,l,w)$, $t_b = (S,L,e',l',w')$ and $t_c = (S,L',e'',l'',w'')$ over nodes $S = \{1,2,3,4\}$,
and edges $L = \{(1,2),(2,3),(3,4)\}$, and $L' = \{(4,3),(3,2),(2,1)\}$.
Furthermore, let $a(r_l)=\{(1,2),(2,3),(3,2),(2,1)\}$ and $a(r_r)=\{(2,3),(3,4),(4,3),(3,2)\}$,
meaning that $r_l$ forms a resource area covering the left and middle part of the graph while $r_r$ forms a resource area covering the middle and right part of the graph.
The resources overlap in the middle on edges $(2,3)$ and $(3,2)$.
Consider train line $t_a$ entering resource area $\{(1,2),(2,3)\}$ via edge $(1,2)$, using resource $r_l$, before train line $t_b$ enters this edge;
it immediately follows that $t_a$ also has to enter edge $(2,3)$ before train line $t_b$.
Given that edge $(2,3)$ is also in the resource area $\{(2,3),(3,4)\}$, it is impossible for train line $t_b$ to overtake $t_a$,
and the sequence of entering these resource areas is the same.
The same holds for train lines $t_a$ and $t_c$.
Once $t_a$ enters resource area $\{(1,2),(2,3)\}$ before $t_c$ enters resource area $\{(3,2),(2,1)\}$,
it follows that $t_a$ also has to enter resource area $\{(2,3),(3,4)\}$ before train line $t_c$ can enter resource area $\{(4,3),(3,2)\}$,
as both trains cannot get past each other because their respective resource areas overlap.

\begin{lstlisting}[label=lst:oo1,caption={Optional overlap constraints (part I).},firstnumber=86]
ra_overlap(T,R,A,R',A') :- |\label{enc:overlap_start}|
   ra(T,R,A,(V,V')), ra(T,R',A',(V,V')), R != R'.|\label{enc:overlap_end}|
:- seq(T,T',R,A,A'), not seq(T,T',R',B,B'),|\label{enc:start_overlapdyn}|
   ra_overlap(T,R,A,R',B), ra_overlap(T',R,A',R',B'),
   1 #sum {1 : route(T,(V,V')),|\label{enc:start_overlapdyn_route}|
               ra(T,R,A,(V,V')),   ra(T,R',B,(V,V')),
               route(T',(U,U')),
               ra(T',R,A',(U,U')), ra(T',R',B',(U,U'))},|\label{enc:end_overlapdyn_route}|
   1 #sum {1 : shared(T,T',R',B,B');|\label{enc:end_overlapdyn_start}|
           1 : decided(T,T',R',B,B');
           1 : decided(T',T,R',B',B)}.|\label{enc:end_overlapdyn_end}|

\end{lstlisting}
%
In Listing~\ref{lst:oo1},
we first compute pairs of resource areas that overlap in lines~\ref{enc:overlap_start} and~\ref{enc:overlap_end}
by deriving atoms \code{ra\_overlap($t$,$r$,$a$,$r'$,$a'$)}
representing that for a train line $t$, resource areas $A\in \mathcal{A}^t_r$ and $A'\in \mathcal{A}^t_{r'}$ share at least one edge for distinct resources $r$ and $r'$.
Note that $a$ and $a'$ are used as identifiers for resource areas $A$ and $A'$, respectively.
The integrity constraint in
lines~\ref{enc:start_overlapdyn}--\ref{enc:end_overlapdyn_end} enforces that train lines routed over an overlap have the same sequence over both pairs of resource areas.
Lines~\ref{enc:start_overlapdyn_route}--\ref{enc:end_overlapdyn_route} require that both train lines visit at least one edge that is shared by the overlapping resource areas.
We ensure that the sequence is only enforced,
if a resource conflict is possible between these two train lines in lines~\ref{enc:end_overlapdyn_start}--\ref{enc:end_overlapdyn_end}.
The predicate \code{decided} is used to extend this constraint to other cases, where we can exploit overlap to reduce the search space.
We examine such a case in the following section.

\subsubsection{Resource overlap II}

We improve the optional overlap constraint from Listing~\ref{lst:oo1} in Listing~\ref{lst:oo2} by additionally considering sequences over resource areas which can be determined before search.

\begin{minipage}{\linewidth}
\begin{lstlisting}[label=lst:oo2,caption={Optional overlap constraints (part II).},firstnumber=97]
decided(T,T',R,A,A') :-|\label{enc:oo2cond1_start}|
   l_ra(T,R,A,L), e_ra(T',R,A',E), T != T', L < E.|\label{enc:oo2cond1_end}|
decided(T,T',R,A,A') :-|\label{enc:start_oo2cond2}|
   ra(T,R,A,(V,V')),   set(T,(V,V')),  l(T,V',L), T != T',
   ra(T',R,A',(U,U')), set(T',(U,U')), e(T',U,E), L < E.|\label{enc:end_oo2cond2}|
seq(T,T',R,A,A')     :- decided(T,T',R,A,A').|\label{enc:decide2seq}|

\end{lstlisting}
\end{minipage}
%
Given a resource $r$, the sequence how two different trains $t=(S,L,e,l,w)$ and $t'=(S',L',e',l',w')$ leave and enter their respective resource areas $A^t_r$ and ${A}^{t'}_r$ is already decided if either
\begin{enumerate}
\item the latest entry time of train line $t$ into resource area $A^t_r$ is before the earliest entry time of train line $t'$ into resource area $A^{t'}_r$
  (cf.~lines~\ref{enc:oo2cond1_start} and~\ref{enc:oo2cond1_end}), or
\item there exist two edges $(v,v')\in A^t_r$ and $(u,u')\in A^{t'}_r$ that are included in every possible path of train line $t$ and $t'$, respectively,
  where the latest entry time $l(v')$ is before the earliest entry time $e'(u)$
  (cf.~lines~\ref{enc:start_oo2cond2}--\ref{enc:end_oo2cond2}).
\end{enumerate}
For the latter case, we use precomputed atoms \code{set($t$,($v$,$v'$))} to denote edges $(v,v')$ that are already set to be in every path of a train line $t$.
Reconsider our previous example with train lines $t_a =(S,L,e,l,w)$ and $t_b = (S,L,e',l',w')$ where $S=\{1,2,3,4\}$ and $L=\{(1,2),(2,3),(3,4)\}$.
Given that $l(3)<e'(1)$, it is already decided that train line $t_a$ leaves resource area $\{(1,2),(2,3)\}$ before $t_b$ enters this resource area.
By deriving this information in Listing~\ref{lst:oo2}, Listing~\ref{lst:oo1} ensures that the same sequence is used for resource area $\{(2,3),(3,4)\}$.

No matter which case holds, we derive the atom \code{decided($t$,$t'$,$r$,$a$,$a'$)} to state that train line $t$ leaves resource area $A^t_r$
before train line $t'$ enters resource area ${A'}^{t'}_r$.
Converting these atoms into sequences in Line~\ref{enc:decide2seq} enhances propagation of the rules in Listing~\ref{lst:oo1},
as these already decided sequences can be used in the integrity constraint in Line~\ref{enc:start_overlapdyn} to derive the truth value of additional sequences.
Note that since the additional instances of the \code{seq/$5$} predicate
serialize resource areas where the time intervals do not overlap,
solutions of the encodings in listings~\ref{lst:conflicts} and \ref{lst:scheduling} remain unchanged.

\subsubsection{Sequence acyclicity}
Whenever more than two train lines have to be serialized on a resource,
the resulting sequence can never be cyclic.
In more detail, given a resource $r$ and three train lines $t_1$, $t_2$, and $t_3$, if $t_1$ leaves resource area $A^{t_1}_r$ before $t_2$ enters resource area $A^{t_2}_r$,
and $t_2$ leaves resource area $A^{t_2}_r$ before $t_3$ enters resource area $A^{t_3}_r$,
it follows that $t_1$ leaves area $A^{t_1}_r$ before $t_3$ enters area $A^{t_3}_r$ as well.
This is guaranteed by the constraints in Condition~\eqref{cond:conflict} and realized by the difference constraint atoms in Listing~\ref{lst:scheduling} lines~\ref{enc:start_seqdiff}--\ref{enc:end_seqdiff}.

\begin{minipage}{\linewidth}
\begin{lstlisting}[label=lst:ot,caption={Optional acyclicity constraints.},firstnumber=103]
:- seq(T,T',R,A,A'), seq(T',T'',R,A',A''),
   shared(T,T'',R,A,A''), not seq(T,T'',R,A,A'').
\end{lstlisting}
\end{minipage}
While this condition would be eventually derived by the difference constraints propagator,
we immediately restrict the search space by providing this explicit constraint, realized in Listing~\ref{lst:ot}.
Furthermore, by expressing this timing-related condition in terms of regular ASP atoms,
we can leverage the propagation mechanism of state-of-the-art ASP solving technology,
which is currently faster and more sophisticated compared to the one used in the difference constraints propagator.

\subsubsection{Sequence heuristic}

\begin{minipage}{\linewidth}
\begin{lstlisting}[label=lst:h1,caption={Heuristic that orders conflicting train lines by their possible arrival times.},firstnumber=105]
#heuristic seq(T,T',R,A,A') :
   shared(T,T',R,A,A'),
   e_ra(T,R,A,E), e_ra(T',R,A',E'),
   l_ra(T,R,A,L), l_ra(T',R,A',L'). [E'-E - (L-L'),sign]

\end{lstlisting}
\end{minipage}
%
The heuristic in Listing~\ref{lst:h1} attempts to order conflicting train lines by their possible arrival times at the areas where the conflict is located.
In essence, we analyze how the time intervals of the train lines are situated and prefer their sequence accordingly.
Given two train lines $t$ and $t'$ with intervals $[e,l]$ and $[e',l']$ at the conflicting areas, respectively,
we calculate $s=e'-e-(l-l')$ to determine whether $t$ should be scheduled before $t'$.
If $s$ is positive, the preferred sign of the sequence atom is also positive, thus preferring $t$ to go before $t'$,
if it is negative, the opposite is expressed.
In detail, $e'-e$ is positive if $t'$ may arrive later than $t$ thus making it more likely that $t$ can go first without delaying $t'$.
Similarly, if $l-l'$ is negative, $t'$ may leave later,
suggesting $t$ to go first.
If the results of both expressions have the same sign,
one interval is contained in the other and if the difference is positive, the center of the interval of $t$ is located earlier than the center of the interval of $t'$.
For example, in Figure~\ref{fig:times},
we see that $t_1$ and $t_2$ share resource $\mathit{sw}_1$ in Node 3 and the time intervals in which they potentially arrive at those edges are $[300,600]$ and $[180,540]$, respectively.
Given that $180-300-(600-540)=-180$, we prefer $t_2$ to be scheduled before $t_1$,
which in the example is clearly the correct decision,
since $t_2$ precedes $t_1$ without delaying $t_1$.


\section{Experiments}\label{sec:experiments}

We evaluate our train scheduling solution using the hybrid ASP system \clingoDL~v1.1, which is built upon the API of \clingo~5.3.%
\footnote{Both are development versions available on \url{https://github.com/potassco/{clingoDL,clingo}/} with commit hashs \texttt{\#f1a185e} and \texttt{\#f8a51134}, respectively.}
As benchmark set, we use 25 real-world instances crafted by domain experts at Swiss Federal Railways (SBB).
The instances capture parts of the railway network between three Swiss cities, namely Z\"urich, Chur and Luzern,
and vary in number of train lines, size and depths of railway network and timing constraints.
The biggest instances contain the whole railway network and up to 467 train lines taken from long distance, regional, suburban and freight traffic
between these three cities.
Thus, we tackle instances with approximately six hours of the full train schedule on a railway network covering approximately 200 km.
For brevity, we omit slight grounding and propagation optimizations in the encoding presented in Section~\ref{sec:approach};
the full encoding and instance set is available at {\url{https://github.com/potassco/train-scheduling-with-hybrid-asp/}}.
We use the best optimization configuration determined in \cite{abjoossctowa19a}.
In detail, we use two threads,
one running with a model-guided optimization strategy iteratively producing models of descending cost until the optimum is found,
and the other with a core-guided optimization strategy~\cite{ankamasc12a} relying on successively identifying and relaxing unsatisfiable cores until a model is obtained.
Furthermore, the delay minimization is modeled by setting $D_{t,v}=\bind{t}{v}\cup \lind{t}{v}{180}$ for each train line $t$ at node $v$,
thus, there are six thresholds in total,
one if there is any delay at all,
and five with a distance of 180 seconds in between with the last one being at 15 minutes.
We use lexicographic optimization minimizing delay first and route penalty second.

As a preliminary step, we benchmark each domain-specific heuristic and optional constraints
individually,
and removed the ones that did not yield any improvement compared to \clingoDL's default setting.
\footnote{In particular, we dropped two heuristics proposed in \cite{abjoossctowa19a} since
  they did not improve solving performance with the new resource area and height-based encoding.}
This might be either due to the fact that the search space was too drastically reduced and as such less impact can be achieved by the branching heuristics,
or that the switch to lexicographic optimization made some heuristics obsolete.

In total, we examined four optional encoding parts, one domain-specific heuristic and three optional constraints, and all their combinations.
More precisely,
we consider the sequence heuristic (\hseq),
resource overlap I (\ooone),
resource overlap II (\ootwo),
and sequence acyclicity (\ot),
and all seven possible combinations
(given that \ooone\ is contained in \ootwo).
We compare the results to \clingoDL's default settings without optional encoding parts and domain-specific heuristic (\none). 

All benchmarks ran on Linux with a Xeon E3-1260L quad-core 2.9~GHz processors and 32~GB RAM;
each run limited to 3 hours and 32GB RAM.
For all our experiments, we asked \clingoDL\ to return one optimal solution,
and we then validate feasibility and quality via an external tool provided by SBB.%
\footnote{\url{https://github.com/potassco/train-scheduling-with-hybrid-asp/}.}

Table~\ref{tab:conf:one} shows the average runtime (\runtime) over all 25 instances,
which includes grounding and solving,
grounding time (\grounding), choices (\choices) and conflicts (\conflicts)
for individual domain-specific heuristics and optional constraints;
their combinations are given in Table~\ref{tab:conf:two}.
The best results in a row are marked bold.
Note that for all configurations and instances, \clingoDL\ was able to find an optimal solution,
thus, no time- or memory-outs blur average time as a means for comparing the performance of the different configurations.
Grounding the largest instances within the time limit is enabled by our preprocessing and the height-based naming scheme for the integer variables,
due to their significant impact on reducing grounding size.

\begin{table}[ht]
\centering
\begin{tabular}{ p{1cm} || rrrr }
\conf & \runtime & \grounding & \choices & \conflicts\\\cline{1-5}
\none   & 550          & \textbf{45} & 242516740          & 9135 \\
\hseq   & 162          & \textbf{45} & 51933286           & 7363 \\
\ooone  & 118          & 53          & 13003796           & 4038 \\
\ootwo  & \textbf{105} & 62          & \textbf{8377540}   & \textbf{3530} \\
\ot     & 435          & 46          & 116873540          & 8755 \\

\end{tabular}
\caption{Average runtime~(\runtime), grounding time~(\grounding), choices~(\choices) and conflicts~(\conflicts) for individual domain-specific heuristics and optional constraints.\label{tab:conf:one}}
\end{table}
%
We clearly see in Table~\ref{tab:conf:one} that \ootwo\ by itself improves solving performance the most,
drastically reducing choices and conflicts compared to \none\
and displaying the lowest average runtime overall.
This is closely followed by \ooone.
The heuristic \hseq\ still clearly improves the solving performance while optional constraint \ot\ only slightly improves upon the default.
We can observe that grounding time is higher for all optional constraints, especially for \ootwo.
This is to be expected as additional rules and integrity constraint are added to the encoding.
The improvement in solving performance through the reduced search space vastly outweighs the decrease in grounding performance though.

\begin{table}[ht]
\begin{tabular}{ l || rrrr }
\conf & \runtime & \grounding & \choices & \conflicts\\\cline{1-5}
\hseq/\ooone     & 72          & 53          & 2877997          & 3097 \\
\hseq/\ootwo     & 77          & 63          & 2097689          & 2606 \\
\hseq/\ot        & 146         & \textbf{46} & 29245280         & 6814 \\
\ooone/\ot       & 99          & 55          & 5934253          & 3905 \\
\ootwo/\ot       & 93          & 64          & 3618248          & 3293 \\
\hseq/\ooone/\ot & \textbf{71} & 55          & 1884225          & 2706 \\
\hseq/\ootwo/\ot & 76          & 64          & \textbf{1486261} & \textbf{2525} \\\cline{1-5}
\vbs             & 68          & 53          & 2383874          & 2560 \\
\end{tabular}
\caption{Average runtime~(\runtime), grounding time~(\grounding), choices~(\choices) and conflicts~(\conflicts) for combinations of domain-specific heuristics and optional constraints.\label{tab:conf:two}}
\end{table}
%
As for the composite configurations in Table~\ref{tab:conf:two},
we observe that all of them are an improvement over the individual heuristics and optional constraints.
The largest combinations also yield the best performance overall.
Here, we see that, while \hseq/\ootwo/\ot\ traverses the search space most effectively (yielding least choices and conflicts),
\hseq/\ooone/\ot\ has a slightly better runtime performance due to lower grounding time.
This shows, first, that the combination \hseq/\ooone/\ot\ alleviate some weaknesses compared to the individual configurations, as \ooone\ alone was slower than \ootwo.
And second, that both composite configurations might prove useful in the future
since for harder instances the improvement in terms of the search might outweigh the decrease in grounding performance.
The last row \vbs\ displays the virtual best solver that averages the best performance regarding runtime among all configurations.
We see that the performance of \vbs\ is close to the best configurations and as such choosing one among \hseq/\ooone/\ot,
\hseq/\ootwo/\ot and \hseq/\ooone\ should be a good choice as default configuration for most instances.

\begin{table}
\centering
\scriptsize
\begin{tabular}{ l || r|rr|rr|rr|rr | rrr }
\instance & \trains & \resources & \subsumed & \resourcextl  & \resourceareas & \edgeconflicts & \areaconflicts & \nodenaming & \toponaming & \runtime & \aquality & \quality \\\cline{1-13}
\textsc{p1}	&4	    &115	&63	    &908	&134	&537	 &36	&318	&294     & 0   & (0,0) & (0,0)\\
\textsc{p2}	&58	    &530	&205	&15180	&3132	&21090	 &2875	&4409	&4371    & 4   & (0,0) & (0,0)\\
\textsc{p3}	&143	&539	&205	&28754	&6082	&64766	 &9158	&8759	&8662    & 7   & (0,0) & (0,0)\\
\textsc{p4}	&148	&553	&207	&34461	&6810	&106749	 &13745	&9425	&9114    & 9   & (0,1) & (0,1)\\
\textsc{p5}	&149	&553	&207	&34482	&6814	&107109	 &13802	&9430	&9119    & 11  & (9,6) & (27,6)\\
\textsc{p6}	&365	&567	&210	&175722	&25021	&912840	 &44908	&38941	&28099   & 53  & (0,0) & (0,0)\\
\textsc{p7}	&467	&567	&210	&232827	&33934	&1591476 &82562	&52060	&38119   & 89  & (0,0) & (0,0)\\
\textsc{p8}	&133	&528	&195	&102456	&12454	&1086507 &34996	&21559	&12277   & 43  & (0,0) & (0,0)\\
\textsc{p9}	&287	&568	&192	&159316	&18862	&4180212 &103225&34488	&18513   & 179 & (0,0) & (0,0)\\
\textsc{p9r} &83	    &505	&188	&77716	&7541	&1998885 &28572	&14579	&6591    & 44  & (0,0) & (0,0)\\ 
\textsc{p9ri}&83	    &505	&188	&77716	&7541	&1316502 &18616	&14579	&6591    & 35  & (0,0) & (0,0)\\ 
\textsc{ts}	&131	&915	&303	&42590	&8026	&263278	 &25036	&12199	&10695   & 16  & (0,0) & (0,0)\\ 
\textsc{tse} &132	&915	&303	&43011	&8105	&274020	 &25934	&12303	&10786   & 18  & ($\approx 0$,0) & (6,0)\\ 
\textsc{tl1}	&447	&920	&301	&126271	&24068	&896199	 &87094	&37163	&32533   & 55  & (0,0) & (0,0)\\ 
\textsc{tl1e}	&448	&920	&301	&126692	&24147	&907059	 &88003	&37267	&32624   & 67  & ($\approx 0$,0) & (11,0)\\ 
\textsc{tl1ei}&448	&920	&301	&126692	&24147	&187393	 &20612	&37267	&32624   & 41  & ($\approx 0$,0) & (11,0)\\ 
\textsc{tl2}&448	&952	&272	&183865	&29531	&494446	 &30961	&51707	&32781   & 116 & ($\approx 0$,0) & ($\approx 0$,0)\\ 
\textsc{tl3}&448	&952	&272	&183865	&29531	&488645	 &30737	&51707	&32781   & 111 & (0,0) & (0,0)\\ 
\textsc{tl4}&448	&940	&271	&165909	&28442	&359927	 &28115	&47595	&32762   & 87  & ($\approx 0$,0) & (1,0)\\ 
\textsc{tl5}&448	&940	&271	&165909	&28442	&357267	 &28045	&47595	&32762   & 86  & (0,0) & (0,0)\\ 
\textsc{tl6}&448	&925	&290	&131691	&25058	&197830	 &21663	&38494	&32752   & 41  & ($\approx 0$,0) & ($\approx 0$,0)\\ 
\textsc{tl7}&448	&925	&290	&131691	&25058	&197148	 &21544	&38494	&32752   & 42  & (0,0) & (0,0)\\
\textsc{tl8}&448	&955	&272	&216830	&31805	&871732	 &40851	&58722	&32781   & 177 & (0,0) & (0,0)\\ 
\textsc{tl9}&448	&955	&272	&216830	&31805	&863326	 &40535	&58722	&32781   & 173 & (0,0) & (0,0)\\ 
\textsc{tl10}&451	&952	&274	&185912	&29687	&1216327 &69972	&52184	&33088   & 177 & ($\approx 0$,75) & (10,75)\\\cline{1-13} 
    &       &       &34\%   &       &83\%   &        &92\%  &       &25\%

\end{tabular}
\caption{Detailed information and best results for all benchmark instances.\label{tab:inst}}
\end{table}
%
In the following, we analyze the 25 real-world instances in detail and highlight the best results that we could achieve using the different configurations of our train scheduling solution.
This information is presented in Table~\ref{tab:inst}.
Row \instance\ contains the names of the 25 instances.
Instance names starting with \textsc{p} are the nine instances originally used in \cite{abjoossctowa19a} that were published by SBB;
they only contain part of the railway network between Z\"urich, Chur and Luzern.
Two instances among them are marked with \textsc{r}
and are reduced versions of the largest instance, viz. \textsc{p9}.
Instances starting with \textsc{t}, on the other hand, cover the whole test area between the three cities.
Furthermore, instance names containing \textsc{s} and \textsc{l} cover a shorter (about two hours) or longer time span (about six hours), respectively,
which is reflected in the amount of train lines.
With our problem formulation, one can express different settings of the train scheduling problem.
Normally, all train lines are considered as new and to be taken into account evenly.
Another problem variation is the rescheduling after adding or changing one extra train line.
This is achieved by closely restricting earliest and latest arrival times around the old schedule for existing and unchanged train lines,
while the extra train line is given more freedom.
Instances containing an \textsc{e} are such rescheduling instances.
Similarly, most instances have a uniform time span between the point after which a train line is delayed and the latest possible arrival time.
Unlike this, for names including \textsc{i}, this time span is different for each train line, i.e., different train lines have different capacities for delay.
Instances of the same class mostly vary in the depths of the railway network, i.e., how many alternative routes are available to each train line,
but also slightly in number of train lines, resources and timing constraints.

The following nine columns analyze structural attributes of the instances,
and how much preprocessing and encoding techniques reduce the instance size and search space.
Column~\trains\ shows the number of train lines for each instance.
Column~\resources\ gives the number of resources,
and Column~\subsumed\ the number of subsumed resources.
Subsumed resources can be observed in all instances and constitute on average 34\%,
which can then be safely removed.
Column~\resourcextl\ sums up resources on each edge in each subgraph of all train lines,
which essentially constitutes the basis for resource conflicts using the edge-based conflict resolution.
Column~\resourceareas\ then shows into how many resource areas those individual edges could be distributed.
The difference from \resourcextl\ to \resourceareas\ is on average 83\%,
thus drastically reducing resource entities that might induce conflicts.
The following columns \edgeconflicts\ and \areaconflicts\
display the number of resource conflicts
based on single edges and resource ares, respectively.
They highlight the benefit of introducing resource areas.
We reduce the number of resource conflicts,
and thus the decisions that have to be made to serialize the train lines,
on average by 92\%.
Finally, \nodenaming\ and \toponaming\ shows the number of integer variables needed using node- and height-based naming scheme, respectively.
We save 25\% of variables on average switching to the height-based naming scheme.

In the last three columns, we present the virtual best results regarding runtime across all configurations presented in Table~\ref{tab:conf:one} and~\ref{tab:conf:two}.
In more detail, we show the runtime per instance in Column~\runtime,
and the approximated and exact quality as a pair of delay penalty in minutes and route penalty in Columns~\aquality\ and Column~\quality, respectively.
The approximated quality amounts to the value returned by the ASP system, which is optimal for all instances,
while the exact quality is returned by a tool of SBB.
We were able to solve all instances with a maximum time of about 3 minutes.
We calculated solutions without any delay for all instances where this was possible.
For instances where this was not the case,
experts at SBB deemed the incurred delay as close to optimal.
Of course, we are not able to prove true optimality in all cases since we use an approximation of the delay function.
Note that, for example for instance \textsc{tl10}, we have a rather high difference between approximated delay and actual delay of about 11 minutes.
This is due to the fact that the ASP system only considers a lower bound on delay.
If the approximated optimal value is for example 3 seconds, this could mean that there are three instances of delay.
The amount of delay though could be as much as $3*179=537$ seconds since the next threshold is at 180 seconds in our configuration of the approximation.
Completely avoiding this drawback of the approximation is currently only possible by introducing a threshold for every second,
which deteriorates solving performance.
We judge our current solution to be a good trade-off between solving performance and quality,
since the quality of the results is sanctioned by Swiss Federal Railways and we are able to solve all real-world instances within minutes.


\section{Discussion}\label{sec:discussion}

At its core, train scheduling is similar to classical scheduling problems that were already tackled by ASP.
Foremost, job shop scheduling~\cite{taillard93a} is also addressed by \clingoDL{} and compared to other hybrid approaches in~\cite{jakaosscscwa17a};
solutions based on SMT, CP and MILP are given in \cite{jalini11a,bopasuvi12a,bapanu12a,lijani12a}.
In fact, job shop scheduling can be seen as a special case of our setting,
in which train paths are known beforehand.
Solutions to this restricted variant using MILP and CP are presented in \cite{olismi00a,rodriguez07a}.
The difference to our setting is twofold:
first, resource conflicts are not known beforehand since we take routing and scheduling simultaneously into account.
Second, our approach encompasses a global view of arbitrary precision, i.e.,
we model all routing and scheduling decisions across hundreds of trains lines down to inner-station conflict resolution
and allow for expressing complex connections that also accommodate cyclic train movements.
Furthermore, using hybrid ASP with difference constraints gives us inherent advantages over pure ASP and MILP.
First, we show in~\cite{gekakaosscwa16a} that ASP is not able to solve most shop scheduling instances
since grounding all integer variables leads to an explosion in problem size.
We avoid this bottleneck by encapsulating scheduling in difference constraints and,
hence, avoid grounding integer variables.
Second, while difference constraints are less expressive than linear constraints in MILP,
they are sufficient for expressing the timing constraint needed for train scheduling while being solvable in polynomial time.
Finally, routing and conflict resolution require Boolean variables and disjunctions for which ASP has effective means.

In this work, we contribute a flexible and holistic ASP-based encoding of the train scheduling problem based on a precise formalization.
On the one hand, since we produce timetables from scratch,
our train scheduling approach can be characterized as tactical scheduling~\cite{tornquist06a}.
On the other hand, without changing encoding or problem formalization,
we can easily accommodate problem variations like rescheduling without loosing performance,
as we show for three instances in our empirical evaluation.

As seen in Section~\ref{sec:experiments}, all instances available to us can be solved within minutes.
These instances represent sections of a test area in Switzerland,
with the largest one capturing a complex, interconnected railway network spanning in length about 200 km
and involving up to 467 train lines with local, regional and cargo railway traffic among them.
This is possible in part due to dedicated preprocessing techniques,
such as resource subsumption and resource areas,
that exploit structural redundancies within the railway network
as well as graph compressing methods,
which drastically reduce the size of the problem representation.
Furthermore, we reduce the search space and leverage advanced propagation mechanisms of our state-of-the-art ASP solving technology
by transferring implicit knowledge about difference constraints into the logic program.
Some of these techniques may also be candidates to improve performance for other scheduling problems like the related job shop scheduling.

To advance our approach, multi-shot solving~\cite{gekakasc17a} can be used to incrementally increase train line intervals or dynamically replan schedules by adding or removing train lines.
While our hybrid ASP approach already tackles some real-world instances that are on the larger side,
scalability, for example, to the entirety of Switzerland is still an issue.
For that purpose, we are looking into decomposition techniques,
so that smaller areas using our approach can be combined to solutions for entire countries.


\bibliographystyle{acmtrans}

\appendix 

\section{Example instance}
\label{appendix:facts}
\begin{lstlisting}[caption={Facts representing example instance (figures \ref{fig:paths} and \ref{fig:times}).}, numbers=none, label=enc:facts]
tl(t1). tl(t2). tl(t3).

edge(t1,1,3).  edge(t1,2,3).   edge(t1,3,5).  edge(t1,5,8).
edge(t1,8,10). edge(t1,10,11). edge(t1,10,12).
edge(t2,10,7). edge(t2,7,4).   edge(t2,4,3).
edge(t3,3,6).  edge(t3,6,9).   edge(t3,9,10). edge(t3,10,12).

e(t1,1,240).  l(t1,1,540).  e(t1,2,240).  l(t1,2,540).
e(t1,3,300).  l(t1,3,600).  e(t1,5,360).  l(t1,5,660).
e(t1,8,420).  l(t1,8,720).  e(t1,10,480). l(t1,10,780).
e(t1,11,540). l(t1,11,840). e(t1,12,540). l(t1,12,840).
start(t1,1).  start(t1,2).  end(t1,11).   end(t1,12).

e(t2,10,0).   l(t2,10,360). e(t2,7,60).  l(t2,7,420).
e(t2,4,120).  l(t2,4,480).  e(t2,3,180). l(t2,3,540).
start(t2,10). end(t2,3).

e(t3,3,180).  l(t3,3,540). e(t3,6,240).  l(t3,6,600).
e(t3,9,300).  l(t3,9,660). e(t3,10,360). l(t3,10,720).
e(t3,12,420). l(t3,12,780).
start(t3,3).  end(t3,12).

potlate(t1,1,451,1).  potlate(t1,2,451,1). penalty((1,3),1).
potlate(t2,10,241,1). potlate(t3,3,421,1).

resource(sw1,(1,3)).   resource(sw1,(2,3)).
resource(sw1,(4,3)).   resource(sw1,(3,5)).
resource(sw1,(3,6)).   resource(sw2,(8,10)).
resource(sw2,(9,10)).  resource(sw2,(10,7)).
resource(sw2,(10,11)). resource(sw2,(10,12)).

connection(1,t2,(4,3),t3,(3,6),0,0,3,3).
free(1,t2,(4,3),t3,(3,6),sw1).
connection(2,t1,(10,11),t3,(10,12),60,#inf,10,12).
connection(3,t1,(10,12),t3,(10,12),60,#inf,10,12).

set(t1,(3,5)).  set(t1,(5,8)). set(t1,(8,10)).
set(t2,(10,7)). set(t2,(7,4)). set(t2,(4,3)).
set(t3,(3,6)).  set(t3,(6,9)). set(t3,(9,10)).
set(t3,(10,12)).

resource(r(1,3),(1,3)). resource(r(2,3),(2,3)).
resource(r(3,5),(3,5)). resource(r(5,8),(5,8)).
resource(r(7,4),(7,4)). resource(r(4,3),(4,3)).
resource(r(3,6),(3,6)). resource(r(6,9),(6,9)).

ra(t1,sw1,0,(1,3)).    ra(t1,r(1,3),0,(1,3)).
ra(t1,sw1,0,(2,3)).    ra(t1,r(2,3),0,(2,3)).
ra(t1,sw1,0,(3,5)).    ra(t1,r(3,5),0,(3,5)).
ra(t1,r(5,8),0,(5,8)). ra(t1,sw2,0,(8,10)).
ra(t1,sw2,0,(10,11)).  ra(t1,sw2,0,(10,12)).
ra(t2,sw2,0,(10,7)).   ra(t2,r(7,4),0,(7,4)).
ra(t2,sw1,0,(4,3)).    ra(t2,r(4,3),0,(4,3)).
ra(t3,sw1,0,(3,6)).    ra(t3,r(3,6),0,(3,6)).
ra(t3,r(6,9),0,(6,9)). ra(t3,sw2,0,(9,10)).
ra(t3,sw2,0,(10,12)).

potlate(t1,3,511,1).  potlate(t1,5,571,1).
potlate(t1,8,631,1).  potlate(t1,10,691,1).
potlate(t1,11,751,1). potlate(t1,12,751,1).
potlate(t2,7,301,1).  potlate(t2,4,361,1).
potlate(t2,3,421,1).  potlate(t3,6,481,1).
potlate(t3,9,541,1).  potlate(t3,10,601,1).
potlate(t3,12,661,1).

l_ra(t1,sw1,0,660).    l_ra(t1,r(1,3),0,600).
l_ra(t1,r(2,3),0,600). l_ra(t1,r(3,5),0,660).
l_ra(t1,r(5,8),0,720). l_ra(t1,sw2,0,840).
l_ra(t2,sw2,0,420).    l_ra(t2,r(7,4),0,480).
l_ra(t2,sw1,0,540).    l_ra(t2,r(4,3),0,540).
l_ra(t3,sw1,0,600).    l_ra(t3,r(3,6),0,600).
l_ra(t3,r(6,9),0,660). l_ra(t3,sw2,0,780).
e_ra(t1,sw1,0,240).    e_ra(t1,r(1,3),0,240).
e_ra(t1,r(2,3),0,240). e_ra(t1,r(3,5),0,300).
e_ra(t1,r(5,8),0,360). e_ra(t1,sw2,0,420).
e_ra(t2,sw2,0,0).      e_ra(t2,r(7,4),0,60).
e_ra(t2,sw1,0,120).    e_ra(t2,r(4,3),0,120). 
e_ra(t3,sw1,0,180).    e_ra(t3,r(3,6),0,180). 
e_ra(t3,r(6,9),0,240). e_ra(t3,sw2,0,300).

b(sw1,60).    b(sw2,60).    b(r(1,3),60). b(r(2,3),60).
b(r(3,5),60). b(r(5,8),60). b(r(7,4),60). b(r(4,3),60).
b(r(3,6),60). b(r(6,9),60).

w(t1,(1,3),0).  w(t1,(2,3),0).   w(t1,(3,5),0).  w(t1,(5,8),0).
w(t1,(8,10),0). w(t1,(10,11),0). w(t1,(10,12),0).
w(t2,(10,7),0). w(t2,(7,4),0).   w(t2,(4,3),0).
w(t3,(3,6),0).  w(t3,(6,9),0).   w(t3,(9,10),0). w(t3,(10,12),0).

m((1,3),60).   m((2,3),60).   m((3,5),60).   m((5,8),60).
m((8,10),60).  m((10,11),60). m((10,12),60). m((10,7),60).
m((7,4),60).   m((4,3),60).   m((3,6),60).   m((6,9),60).
m((9,10),60).
\end{lstlisting}

\end{document}